\documentclass[conference]{IEEEtran}
\usepackage{times}
\let\labelindent\relax

\usepackage[numbers,sort&compress]{natbib}
\usepackage[shortlabels]{enumitem}
\usepackage{multicol}
\usepackage[utf8]{inputenc}
\usepackage{graphics} 
\usepackage{graphicx} 
\usepackage{times} 
\usepackage{amsmath} 
\usepackage{amssymb}
\usepackage{bm}
\usepackage{amsthm}
\usepackage{array}
\usepackage{soul}
\usepackage{xspace}
\usepackage[bottom]{footmisc}
\usepackage[dvipsnames]{xcolor}
\usepackage{booktabs}       
\usepackage{tabu}
\usepackage{array}
\usepackage[english]{babel}
\usepackage{xcolor}
\usepackage{xspace}
\usepackage{algorithm}
\usepackage{algorithmic}
\usepackage{romannum}
\usepackage{newtxmath}
\usepackage{newtxtext}

\usepackage{amsmath,amsfonts,bm}

\def\eqref#1{equation~\ref{#1}}
\def\Eqref#1{Equation~\ref{#1}}

\def\1{\bm{1}}

\def\vmu{{\bm{\mu}}}
\def\vtheta{{\bm{\theta}}}
\def\va{{\bm{a}}}

\def\vo{{\bm{o}}}

\def\vu{{\bm{u}}}

\def\vw{{\bm{w}}}
\def\vx{{\bm{x}}}
\def\vy{{\bm{y}}}
\def\vz{{\bm{z}}}

\def\mB{{\bm{B}}}

\def\mL{{\bm{L}}}

\def\mQ{{\bm{Q}}}

\DeclareMathAlphabet{\mathsfit}{\encodingdefault}{\sfdefault}{m}{sl}
\SetMathAlphabet{\mathsfit}{bold}{\encodingdefault}{\sfdefault}{bx}{n}

\DeclareMathOperator*{\argmax}{arg\,max}
\DeclareMathOperator*{\argmin}{arg\,min}

\DeclareMathOperator{\sign}{sign}

\usepackage[bookmarks=true]{hyperref}
\usepackage{url}

\theoremstyle{plain}
\newtheorem{theorem}{Theorem}
\newtheorem*{theorem*}{Theorem}

\setlist[enumerate]{nosep}

\def\vxi{{\bm{\xi}}}
\def\vmu{{\bm{\mu}}}
\def\vnu{{\bm{\nu}}}
\def\vDelta{{\bm{\Delta}}}
\def\vlambda{{\bm{\lambda}}}

\usepackage{etoolbox}
\makeatletter
\patchcmd{\@makecaption}
{\scshape}
{}
{}
{}
\makeatother

\begin{document}
\title{Robust Value Iteration for Continuous Control Tasks}

\author{%
\authorblockN{
Michael Lutter$^{1,2}$, %
Shie Mannor$^{1,3}$, %
Jan Peters$^{2}$, %
Dieter Fox$^{1,4}$, %
Animesh Garg$^{1,5}$}%
\authorblockA{
$^{1}$Nvidia, $^{2}$TU Darmstadt, $^{3}$ Technion, $^{4}$ University of Washington, $^{5}$University of Toronto \& Vector Institute}
}

\maketitle
\IEEEpeerreviewmaketitle

\begin{abstract}
When transferring a control policy from simulation to a physical system, the policy needs to be robust to variations in the dynamics to perform well. Commonly, the optimal policy overfits to the approximate model and the corresponding state-distribution, often resulting in failure to trasnfer underlying distributional shifts. 
%
In this paper, we present Robust Fitted Value Iteration, which uses dynamic programming to compute the optimal value function on the compact state domain and incorporates adversarial perturbations of the system dynamics. The adversarial perturbations encourage a optimal policy that is robust to changes in the dynamics. Utilizing the continuous-time perspective of reinforcement learning, we derive the optimal perturbations for the states, actions, observations and model parameters in closed-form. Notably, the resulting algorithm does not require discretization of states or actions. Therefore, the optimal adversarial perturbations can be efficiently incorporated in the min-max value function update.
%
We apply the resulting algorithm to the physical Furuta pendulum and cartpole. By changing the masses of the systems we evaluate the quantitative and qualitative performance across different model parameters. We show that robust value iteration is more robust compared to deep reinforcement learning algorithm and the non-robust version of the algorithm. 
Videos of the experiments are shown at \textcolor{blue}{\url{https://sites.google.com/view/rfvi}}
\end{abstract}

\section{Introduction}
\noindent 
To avoid the laborious and potentially dangerous training of control policies on the physical system, Simulation to reality transfer (sim2real) learns a policy in simulation and evaluates the policy on the physical system. When transferred to the real world, the policy should solve the task and obtain a comparable reward to the simulation.  
Therefore, the goal of sim2real is to learn a policy that is robust to changes in the dynamics and successfully bridges the simulation-reality gap. Naive reinforcement learning (RL) methods usually do not succeed for sim2real as the resulting policies overfit to the approximate simulation model. Therefore, the resulting policies are not robust to changes in the dynamics and fail to solve the task in the real world. In contrast, sim2real methods extending RL with domain randomization~\cite{muratore2018domain, muratore2021data, chebotar2019closing, ramos2019bayessim} or adversarial disturbances~\cite{bansal2017hamilton, isaacs1999differential, pinto2017robust, mandlekar2017adversarially} have shown the successful transfer to the physical world~\cite{xie2021dynamics}.

\begin{figure}[t]
    \centering
    \includegraphics[width=\columnwidth]{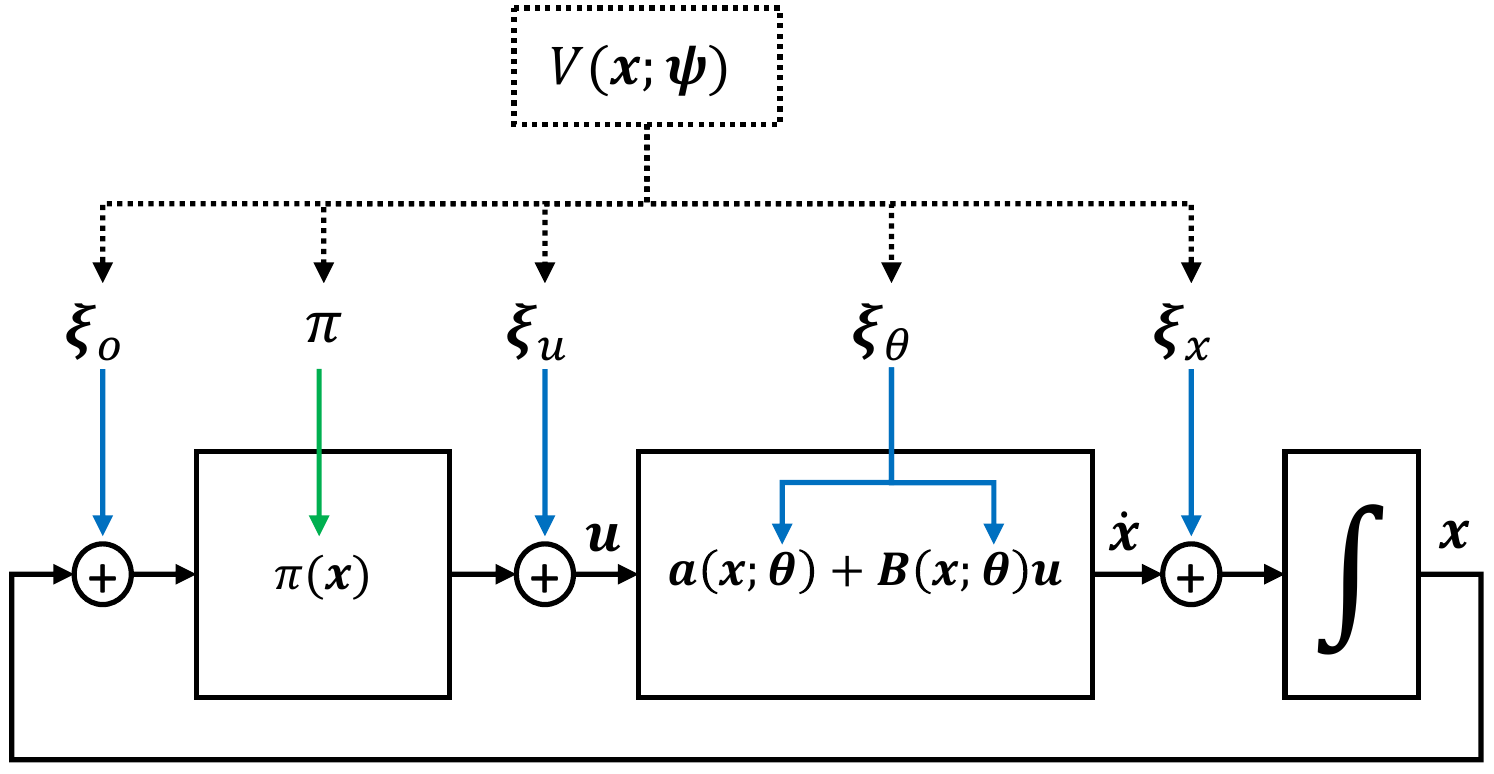}
    \vspace{-2.em}
    \caption{The control chart of robust fitted value iteration (rFVI) for continuous states, actions and time. The deterministic optimal policy and deterministic adversaries, which add an bias to the system dynamics, only depend on shared value function. While the optimal policy performs hill-climbing the adversaries perform steepest descent following the value function gradient.}
    \label{fig:control_flow}
    \vspace{-1.em}
\end{figure} 

\begin{figure*}[t]
    \centering
    \includegraphics[width=\textwidth]{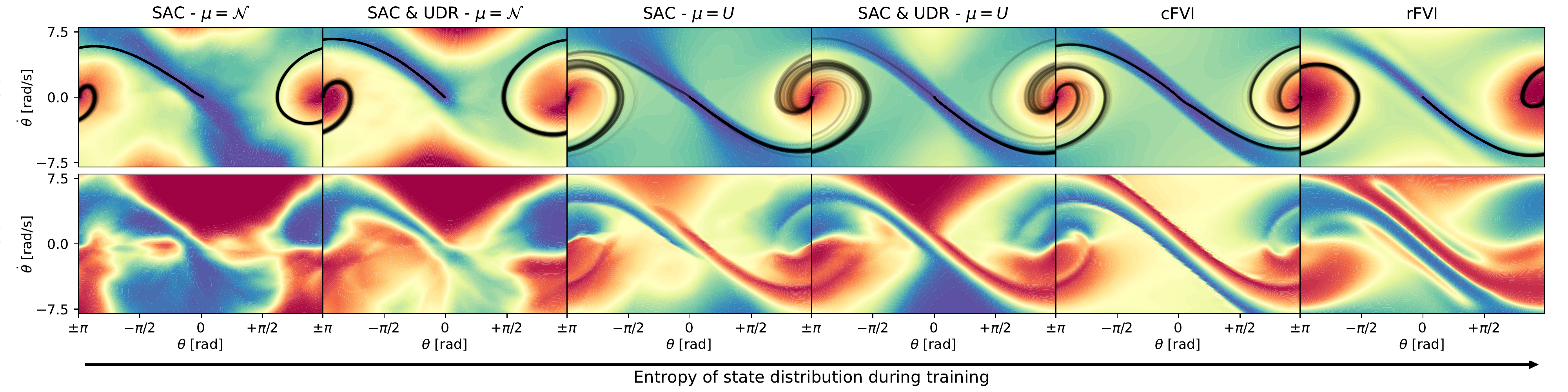}
    \vspace{-2.em}
    \caption{The optimal Value function $V^{*}$ and policy $\pi^{*}$ of rFVI, cFVI and four different variations of SAC. All policies achieve nearly identical reward on the nominal dynamics model. The variations of SAC demonstrate the change of the policy when increasing the entropy of the state distribution during training. The entropy is increased by enlarging the initial state distribution $\mu$ and using domain randomization. For SAC and $\mu = \mathcal{N}(\pm \pi, \sigma)$ the optimal policy is only valid on the optimal trajectory. For SAC UDR and $\mu = \mathcal{U}(-\pi, +pi)$, the policy is applicable on the complete state domain. rFVI and cFVI perform value iteration on the compact state domain and naturally obtain an optimal policy applicable on the complete state-domain. rFVI adapts $V^{*}$ and $\pi^{*}$ to have a smaller ridge leading up to the upright pendulum and exerts higher actions when deviating from the optimal trajectory.}
    \label{fig:value_fun_pendulum}
    \vspace{-1.em}
\end{figure*} 

\medskip
\noindent
In this paper, we focus on adversarial disturbances to bridge the simulation-reality gap. In this paradigm the RL problem is formulated as a two player zero-sum game~\cite{isaacs1999differential}. The optimal policy wants to maximize the reward while the adversary wants to minimize the reward. For control tasks, the policy controls the system input $\vu$ and the adversary $\vxi$ controls the dynamics. For example, the adversary can change the state, action, observation, model parameters or all of them within limits. Therefore, this problem formulation optimizes the worst-case performance and not the expected reward as standard RL. The resulting policy is robust to changes in the dynamics as the planning in simulation uses the worst-case approximation which includes the physical system \cite{garcia2015comprehensive}. 

\medskip \noindent
In this adversarial RL setting, \smallskip
\begin{enumerate} [wide=0pt]
    \item we show that the optimal action and optimal adversary can be directly computed using the value function estimate if the reward is separable into state and action reward and the continuous-time dynamics are non-linear and control-affine. We derive the solution for the state, action, observation and model bias analytically. Therefore, this paper extends the existing analytic solutions from continuous-time RL \cite{lyshevski1998optimal, doya2000reinforcement, morimoto2005robust, lutter2019hjb}. \vspace{0.25em}
    \item we propose robust fitted value iteration~(rFVI). This algorithm solves the adversarial RL problem with continuous states, actions and disturbances by leveraging the analytic expressions and value iteration. Using this approach, the continuous states and actions do not need to be discretized as in classical methods \cite{bellman1957dynamic, puterman1994markov, sutton1998introduction} or require multiple optimizations as the modern actor-critic approaches \cite{pinto2017robust, pinto2017supervision}. \vspace{0.25em}
    \item we provide an in-depth evaluation of rFVI on the physical system. We benchmark this algorithm on the real-world Furuta pendulum and cartpole. To test the robustness, we perturb the model parameters by applying additional weights. The performance is compared to standard deep RL algorithms with and without domain randomization.
\end{enumerate}

\medskip
\noindent
Therefore the contributions of this paper are the derivation of the analytic adversarial actions, the introduction of robust fitted value iteration and the extensive evaluation on multiple physical systems. In the evaluation we focus on two under-actuated systems to provide an in-depth qualitative and quantitative analysis of the various algorithms in the physical world. 

\bigskip
\noindent
In the following we introduce the continuous-time formulation of adversarial RL (Section \ref{sec:problem}).  Section \ref{sec:rFVI} derives the optimal adversaries and introduces robust fitted value iteration. Finally, we describe the experimental setup and report the performance of the algorithms on the physical system in Section \ref{sec:experiments}.      

\section{Problem Statement} \label{sec:problem}
\noindent
The infinite horizon continuous-time RL optimization with the adversary $\vxi$ is described by
\begin{gather}
\pi^*(\vx) = \argmax_{\pi} \: \inf_{\bm{\xi} \in \Omega} \: \int_{0}^{\infty} \exp(-\rho t) \:\: r_{c}(\vx_t, \vu_t) \: dt,
\label{eq:cont_policy} \\ 
V^{*}(\vx) = \max_{\vu} \: \inf_{\bm{\xi} \in \Omega} \: \int_{0}^{\infty} \exp(-\rho t) \: r_c(\vx_{t}, \vu_{t}) \:  dt,
\label{eq:cont_val} \\
\vx(t) = \vx_0 + \int_{0}^{t} f_c(\vx_{\tau}, \vu_{\tau}, \: \bm{\xi}_{\tau}) \: d\tau, 
\end{gather}
with the state $\vx$, action $\vu$, admissible set $\Omega$, discounting constant $\rho \in (0, \infty]$, reward $r_c$, dynamics~$f_c$, optimal value function $V^{*}$ and policy $\pi^{*}$ \cite{isaacs1999differential, morimoto2005robust}. The order of the optimizations can be switched as the optimal actions and disturbance remain identical \cite{isaacs1999differential}. 
The adversary $\bm{\xi}$ must be constrained to be in the set of admissible disturbances $\Omega$ as otherwise the adversary is too powerful and would prevent the policy from learning the task. 

\medskip
\noindent
The deterministic continuous-time dynamics~$f_c$ is assumed to be non-linear w.r.t.~the system state~$\vx$ but affine w.r.t.~the action~$\vu$. Such dynamics are described by
\begin{align}
\dot{\vx} = \va(\vx; \vtheta) + \mB(\vx; \vtheta) \vu \label{eq:affine_dyn}, 
\end{align}
with the non-linear drift $\va$, the non-linear control matrix $\mB$ and the system parameters $\vtheta$. We assume that the approximate parameters $\hat{\vtheta}$ and the approximate equation of motions $\hat{\va}$, $\hat{\mB}$ are known. For most rigid-body systems this assumptions is feasible as the equations of motions can be derived analytically and the system parameters can be measured. The resulting model is only approximate due to the idealized assumption of rigid bodies, measurement error of the system parameters and neglecting the actuator dynamics.    

\begin{table*}[!t]
\centering
\renewcommand{\arraystretch}{1.75}
\setlength\tabcolsep{10pt}
\caption{
    The optimal actions $\vu^k$ and adversarial actions $\xi^{k}$ for the state-, action-, model- and observation bias with the admissible set $\Omega$. \vspace{-10pt}
}
\begin{tabular*}{\textwidth}{l c c c c}
    \toprule
    & State Perturbation
    & Action Perturbation
    & Model Perturbation
    & Observation Perturbation \\
    \midrule
    Dynamics $f_c(\vx, \vu, \xi)$
    & $\va(\vx) + \mB(\vx) \vu  + \xi$
    & $\va(\vx) + \mB(\vx) (\vu  + \xi)$
    & $\va(\theta + \xi) + \mB(\theta + \xi) \vu$
    & $\va(\vx + \xi) + \mB(\vx + \xi) \vu$ \\
    Optimal Action $\vu^{k}$
    & $\nabla \tilde{g}(\mB(\vx)^T \nabla_x V^{k})$
    & $\nabla \tilde{g}(\mB(\vx)^T \nabla_x V^{k})$
    & $\nabla \tilde{g}(\mB(\vx)^T \nabla_x V^{k})$
    & $\nabla \tilde{g}(\mB(\vx)^T \nabla_x V^{k})$ \\
    Optimal Disturbance $\xi^{k}$
    & $-h_{\Omega} \left( \nabla_x V^{k} \right)$
    & $-h_{\Omega} \left( \mB^T \nabla_x V^{k}  \right)$
    & $-h_{\Omega} \left( \left(\frac{\partial \va}{\partial \vx} + \frac{\partial \mB}{\partial \theta} \vu^{k} \right)^T \nabla_x V^{k}  \right)$
    & $-h_{\Omega} \left( \left(\frac{\partial \va}{\partial \vx} + \frac{\partial \mB}{\partial \vx} \vu^{k} \right)^T \nabla_x V^{k} \right)$ \\
\bottomrule
\end{tabular*}
\label{table:adversarial_disturbances}
\end{table*}

\medskip \noindent
The optimal policy and adversary is modeled as a stationary and Markovian policy that applies a state-dependent disturbance. In this case, the worst-case action is deterministic if the dynamics are deterministic. If the adversary would be stochastic, the optimal policies are non-stationary and non-Markovian \cite{zhang2020robust}. This assumption is used in most of the existing literature on adversarial RL \cite{morimoto2005robust, heger1994consideration, zhang2020robust, mandlekar2017adversarially}. 
We consider four different adversaries that alter the state~\cite{heger1994consideration, littman1994markov, nilim2005robust}, action~\cite{morimoto2005robust, pinto2017robust, pinto2017supervision, tessler2019action, pattanaik2017robust, gleave2019adversarial}, observation~\cite{mandlekar2017adversarially, zhang2020robust} and model parameters~\cite{mandlekar2017adversarially}. The different adversaries address potential causes of the simulation gap. The state adversary $\vxi_{x}$ incorporates unmodeled physical phenomena in the simulation. The action adversary $\vxi_{u}$ addresses the non-ideal actuators. The observation adversary $\vxi_{o}$ introduces the non-ideal observations caused by sensors. The model adversary $\vxi_{\theta}$ introduces a bias to the system parameters. All adversaries could be subsumed via a single adversary with large admissible set. However, the resulting dynamics would not capture underlying structure of the simulation gap~\cite{mandlekar2017adversarially} and the optimal policy would be too conservative~\cite{xu2012robustness}. Therefore, we disambiguate between the different adversaries to capture this structure. 
Mathematically the models are described by
\begin{alignat}{1}
\text{State} \hspace{5pt} \vxi_{x}: \hspace{10pt} \dot{\vx} &= \va(\vx; \vtheta) + \mB(\vx; \vtheta) \vu  + \vxi_{\vx} \label{eq:dyn_adv_state}, \\
\text{Action} \hspace{5pt} \vxi_{u}: \hspace{10pt} \dot{\vx} &= \va(\vx; \vtheta) + \mB(\vx; \vtheta) \: \left(\vu + \vxi_{\vu}\right) \label{eq:dyn_adv_action}, \\
\text{Observation} \hspace{5pt} \vxi_{o}: \hspace{10pt} \dot{\vx} &= \va(\vx + \vxi_{\vo}; \vtheta) + \mB(\vx + \vxi_{\vo}; \vtheta) \: \vu  \label{eq:dyn_adv_obs},\\ 
\text{Model} \hspace{5pt} \vxi_{\theta}: \hspace{10pt} \dot{\vx} &= \va(\vx; \vtheta + \vxi_{\theta}) + \mB(\vx; \vtheta + \vxi_{\theta}) \: \vu  \label{eq:dyn_adv_model}.
\end{alignat}
Instead of disturbing the observation, \Eqref{eq:dyn_adv_obs} disturbs the simulation state of the drift and control matrix. This disturbance is identical to changing the observed system state. 

\medskip \noindent
The deterministic perturbation is in contrast to the standard RL approaches, which describe $\vxi_{i}$ as a stochastic variable. However, this difference is due to the worst-case perspective, where one always selects the worst sample at each state. This worst case sample is deterministic if the policies are stationary and Markovian. The filtering approaches of state-estimation are not applicable to this problem formulation as these approaches cannot infer a state-dependent bias. 

\medskip
\noindent
The reward is separable into state reward $q_c$ and action reward $g_c$ described by
\begin{align}
    r_c(\vx, \vu) = q_c(\vx) - g_c(\vu) \label{eq:seperable_rwd}.
\end{align}
The action cost is non-linear, positive definite and strictly convex. The assumptions on the action cost are not limiting as these resulting properties are desirable. The convexity of the action cost enforces that the optimal action is unique. The positive definiteness of $g_c$ penalizes non-zero actions, which prevents the bang-bang controller to be optimal.

\section{Robust Fitted Value Iteration} \label{sec:rFVI}
\noindent
In the following, we summarize continuous fitted value iteration~(cFVI)~\cite{lutter2021cfvi}. Afterwards, we derive the analytic solutions for the optimal adversary and present robust fitted value iteration~(rFVI). This algorithm solves the adversarial RL problem and obtains a robust optimal policy. In contrast, cFVI only solves the deterministic RL problem and obtains an optimal policy that overfits to the approximate dynamics model. 

\subsection{Preliminaries - Continuous Fitted Value Iteration}

\noindent
Continuous fitted value iteration (cFVI)~\cite{lutter2021cfvi} extends the classical value iteration approach to compute the optimal value function for continuous action and state spaces. By showing that the optimal policy can be computed analytically, the value iteration update can be computed efficiently. For non-linear control-affine dynamics (\Eqref{eq:affine_dyn}) and separable reward (\Eqref{eq:seperable_rwd}), the optimal action is described by
\begin{align}
\pi^{k}(\vx) = \nabla \tilde{g}_c \left(\mB(\vx)^{T} \nabla_x V^{k}\right) \label{eq:opt_policy}
\end{align}
with current value function $\nabla_x V^{k}$ and the convex conjugate of the action cost $\tilde{g}_{c}$ \cite{lyshevski1998optimal, doya2000reinforcement, lutter2019hjb}. The convex conjugate is defined as $\nabla \tilde{g}(\vw) = \left[\nabla g(\vw) \right]^{-1}$. For a quadratic action cost, $\nabla \tilde{g}$ is a linear transformation. For barrier shaped action cost, $\nabla \tilde{g}$ re-scales and limits the action range. This solution is intuitive as the optimal policy performs hill climbing on the value function manifold and action cost determines the step size.

\medskip
\noindent
Substituting the analytic policy into the value iteration update, the classical algorithm can be extended to continuous actions, i.e.,
\begin{align}
V^{k+1}_{\text{tar}} &= \max_{\vu} \: r(\vx, \vu) + \gamma V^{k}(f(\vx_{t}, \vu))  \label{eq:val_update}  \\
&= \: r\left(\vx_{t},  \pi^{k}(\vx_t) \right) + \gamma V^{k}\left(f\left(\vx_{t}, \pi^{k}(\vx_t) \right)\right).
\end{align}
Therefore, the closed-form policy enables the efficient computation of the target. Combined with function approximation for the value function~\cite{ernst2005tree, massoud2009regularized, riedmiller2005neural, mnih2015human}, classical value iteration can be extended to continuous state and action spaces without discretization. The fitting of the value function is described by, 
\begin{gather}
\psi_{k+1} = \argmin_{\psi} \sum_{\vx} \| V^{k+1}_{\text{tar}}(\vx) - V(\vx; \:\psi)  \|_{p}^{p}, \label{eq:val_fitting}
\end{gather}
with $\| \cdot \|_p$ being the $\ell_p$ norm. Iterating between computing $V_{\text{tar}}$ and fitting the value function, learns the optimal value function $V^{*}$ and policy $\pi^{*}$. 

\vspace{-0.5em}
\medskip
\subsection{Deriving the Optimal Disturbances}

\noindent 
For the adversarial RL formulation, the value function target contains a max-min optimization described by
\begin{align}
V_{\text{tar}}^{k+1}(\vx) &= \max_{\vu} \: \inf_{\vxi \in \: \Omega} \: r(\vx, \vu) + \gamma V^{k}\big(f(\vx, \vu, \: \vxi)\big). 
\label{eq:val_adv_update}
\end{align}
To efficiently obtain the value function update, the optimal action and the optimal disturbance need to be computed in closed form. We show that this optimization problem can be solved analytically for the described dynamics and disturbance models (Section \ref{sec:problem}). Therefore, the adversarial RL problem can be solved by value iteration.

\medskip\noindent
The resulting optimal actions $\vu^{*}$ and disturbances $\vxi^{*}_{i}$ have a coherent intuitive interpretation. The optimal actions perform steepest ascent by following the gradient of the value function~$\nabla_x V$. The optimal perturbations perform steepest descent by following the negative gradient of the value function. The step-size of policy and adversary is determined by the action cost $g$ or the admissible set $\Omega$. The optimal policy and the optimal adversary is described by 
\begin{align}
\vu^{k} = \nabla \tilde{g} \left(\frac{\partial f_c(.)}{\partial \vu}^{T} \nabla_x V^{k} \right), 
\vxi^{k}_i = - h_{\Omega} \left(\frac{\partial f_c(.)}{\partial \vxi_{i}}^{T} \nabla_x V^{k}\right). \label{eq:opt_adv}
\end{align}
In the following we abbreviate $\left[\partial f_c(.) / \partial \vy\right]^{T} \nabla_{x} V$, as $\vz_{y}$.
For the adversarial policy, $h_{\Omega}$ rescales $\vz_{\xi}$ to be on the boundary of the admissible set. If the admissible set bounds the signal energy to be smaller than $\alpha$, the disturbance is rescaled to have the length $\alpha$. Therefore, the adversary is described by
\begin{align}
\Omega_{E} = \{ \vxi \in \mathbb{R}^{n} \: | \:  \| \vxi \|_{2} \leq \alpha \} \hspace{5pt} \Rightarrow \hspace{5pt} h_{E}(\vz_{\xi}) = \alpha \: \frac{\vz_{\xi}}{\| \vz_{\xi} \|_{2}}. \label{eq:signal_energy}    
\end{align}
If the amplitude of the disturbance is bounded, the disturbance performs bang-bang control. In this case the adversarial policy is described by
\begin{equation}
\begin{aligned}
\Omega_{A} &= \{ \vxi \in \mathbb{R}^{n} \: | \:  \bm{\nu}_{\text{min}} \leq \vxi \leq \bm{\nu}_{\text{max}} \} \\
&\hspace{65pt}\Rightarrow\hspace{5pt} h_{A}(\vz_{\xi}) = \vDelta \sign\left( \vz_{\xi} \right) + \vmu\label{eq:amplitude},
\end{aligned}
\end{equation}
with $\vmu = \left( \vnu_{\text{max}} + \vnu_{\text{min}} \right) / 2$ and $\vDelta = \left( \vnu_{\text{max}} - \vnu_{\text{min}} \right) / 2$.  

\medskip
\noindent
The following theorems derive \Eqref{eq:opt_adv}, \ref{eq:signal_energy} and \ref{eq:amplitude} for the optimal policy and the different disturbances. Theorem~1 describes the state adversary, Theorem~2 describes action adversary, Theorem~3 the observation adversary and Theorem~4 describes the model adversary. Following the theorems, we provide sketches of the proofs for the state and model disturbance. The remaining proofs are analogous. The complete proofs for all theorems are provided in the appendix. All solutions are summarized in Table \ref{table:adversarial_disturbances}.

\begin{theorem}
For the adversarial state disturbance (\Eqref{eq:dyn_adv_state}) with bounded signal energy (\Eqref{eq:signal_energy}), the optimal continuous-time policy $\pi$ and state disturbance $\vxi_{x}$ is described by
\begin{align*}
\pi(\vx) &= \nabla \tilde{g} \left(\mB(\vx)^{T} \nabla_x V\right), & \vxi_x &= - \alpha \frac{\nabla_x V}{\| \nabla_x V \|_2}.
\end{align*}
\end{theorem}

\begin{theorem}
For the adversarial action disturbance (\Eqref{eq:dyn_adv_action}) with bounded signal energy (\Eqref{eq:signal_energy}), the optimal continuous-time policy $\pi$ and action disturbance $\vxi_{u}$ is described by
\begin{align*}
\pi(\vx) &= \nabla \tilde{g} \left(\mB(\vx)^{T} \nabla_x V\right), & \vxi_{u} &= - \alpha \frac{\mB(\vx)^{T} \nabla_x V}{\| \mB(\vx)^{T} \nabla_x V \|_2}.
\end{align*}
\end{theorem}

\medskip
\noindent 
\textbf{Proof Sketch Theorem 1} For the admissible set $\Omega_{E}$, \Eqref{eq:val_adv_update} can be written with the explicit constraint. 
This optimization is described by
\begin{gather*}
V_{\text{tar}} = \max_{\vu} \: \min_{\vxi_x} \: r(\vx, \vu) + \gamma V\big(f(\vx, \vu, \: \vxi_x)\big) 
\hspace{7pt} 
\text{s.t.} \hspace{7pt}  \vxi_{x}^{T} \vxi_{x} \leq \alpha^{2}. 
\end{gather*}
Substituting the Taylor expansion for $V(\vx_{t+1})$, the dynamics model and the reward, the optimization is described by
\begin{align*}
V_{\text{tar}} &= \max_{\vu} \: \min_{\vxi_x} \: r + \gamma V + \gamma \nabla_x V^{T} f_c \Delta t + \gamma \mathcal{O}(\vx, \vu, \Delta t) \Delta t 
\end{align*}
with the higher order terms $\mathcal{O}(\vx, \vu, \Delta t)$. In the continuous-time limit, the higher-order terms and the discounting disappear, i.e., $\lim_{\Delta t\rightarrow 0} \mathcal{O}(\vx, \vu, \Delta t) {=} 0$ and $\lim_{\Delta t\rightarrow 0} \exp(-\rho \Delta t) {=} 1$. Therefore, the optimal action is described by 
\begin{align*}
\vu_{t} = \argmax_{\vu} \:  \nabla_x V^{T} \mB \: \vu - g_{c}(\vu) 
\hspace{5pt} \Rightarrow \hspace{5pt}
\vu_{t} = \nabla \tilde{g}_c\big(\mB^{T} \nabla_x V\big).
\end{align*}
The optimal state disturbance is described by
\begin{align*}
\vxi^{*}_x = \argmin_{\vxi_x} \: \nabla_x V^{T} \vxi_x \hspace{10pt} \text{s.t.} \hspace{10pt} \frac{1}{2} \left[ \vxi_{x}^{T} \vxi_{x} - \alpha^{2} \right] \leq 0.
\end{align*}
This constrained optimization can be solved using the Karush-Kuhn-Tucker (KKT) conditions. The resulting optimal adversarial state perturbation is described by
\begin{align*}
\vxi_{x} = - \alpha \frac{\nabla_x V}{\| \nabla_x V \|_2}.
\end{align*}
\hspace{\fill}\qed

\medskip
\begin{theorem}
For the adversarial model disturbance (\Eqref{eq:dyn_adv_model}) with element-wise bounded amplitude (\Eqref{eq:amplitude}), smooth drift and control matrix (i.e., $\va, \mB \in C^{1}$) and $\mB(\vtheta + \vxi_{\theta}) \approx \mB(\vtheta)$, the optimal continuous-time policy $\pi$ and model disturbance~$\vxi_{\theta}$ is described by
\begin{gather*}
\pi(\vx) = \nabla \tilde{g} \left(\mB(\vx)^{T} \nabla_x V\right), \hspace{20pt}
\vxi_{\theta} = -\vDelta_{\nu} \sign \left( \vz_{\vtheta }\right) + \vmu_{\nu} \\
\text{with} \hspace{5pt} \vz_{\theta} = \left(\frac{\partial \va(\vx;\: \vtheta)}{\partial \vtheta} + \frac{\partial \mB(\vx; \: \vtheta)}{\partial \vtheta} \pi(\vx) \right)^T \nabla_x V,
\end{gather*}
parameter mean $\vmu_{\vnu} = \left( \vnu_{\text{max}} + \vnu_{\text{min}} \right) / 2$ and parameter range $\vDelta_{\vnu} = \left( \vnu_{\text{max}} - \vnu_{\text{min}} \right) / 2$.
\end{theorem}

\medskip
\begin{theorem}
For the adversarial observation disturbance (\Eqref{eq:dyn_adv_obs}) with bounded signal energy (\Eqref{eq:signal_energy}), smooth drift and control matrix (i.e., $\va, \mB \in C^{1}$) and $\mB(\vx + \vxi_{o}) \approx \mB(\vx)$, the optimal continuous-time policy $\pi$ and observation disturbance~$\vxi_{o}$ is described by
\begin{gather*}
\pi(\vx) = \nabla \tilde{g} \left(\mB(\vx)^{T} \nabla_x V\right), \hspace{30pt}
\vxi_{o} = - \alpha \frac{\vz_{o}}{\| \vz_{o} \|_2} \\
\text{with} \hspace{5pt} \vz_{o} = \left( \frac{\partial \va(\vx; \: \vtheta)}{\partial \vx} + \frac{\partial \mB(\vx; \: \vtheta)}{\partial \vx} \pi(\vx) \right)^T \nabla_x V.
\end{gather*}
\end{theorem}

\begin{algorithm}[t]
\caption{Robust Fitted Value Iteration (rFVI)}
\label{alg:rFVI}
\begin{algorithmic}
\STATE {\bfseries Input:} Model $f_{c}(\vx, \vu)$, Dataset $\mathcal{D}$ \& Admissible Set $\Omega_{\xi}$
\STATE {\bfseries Result:} Value Function $V^{*}(\vx;\: \psi^{*})$
\WHILE{not converged}
\STATE // Compute Value Target for $\vx \in \mathcal{D}$:\;
\STATE $\vx_{\tau} = \vx_i + \int_{0}^{\tau} f_c(\vx_{t}, \vu_{t}, \vxi^{x}_{t}, \vxi^{u}_{t}, \vxi^{o}_{t}, \vxi^{\theta}_{t}) dt$
\STATE $R_t = \int_{0}^{t} \exp(-\rho \tau) \: r_c(\vx_{\tau}, \vu_{\tau}) d\tau + \exp(-\rho t) V^{k}(\vx_t)$
\STATE $V_{\text{tar}}(\vx_i) = \int_{0}^{T} \beta \: \exp(-\beta t) \: R_t \: dt + \exp(-\beta T) R_T$
\STATE
\STATE // Fit Value Function:
\STATE $\psi_{k+1} = \argmin_{\psi} \sum_{\vx \in \mathcal{D}} \| V_{\text{tar}}(\vx) - V(\vx; \psi) \|^{p}$ \;
\STATE
\IF{RTDP rFVI}
\STATE // Add samples from $\pi^{k+1}$ to FIFO buffer $\mathcal{D}$
\STATE $\mathcal{D}^{k+1} = h(\mathcal{D}^{k}, \{\vx^{k+1}_0 \: \dots \: \vx^{k+1}_N \})$
\ENDIF
\ENDWHILE
\end{algorithmic}
\end{algorithm}

\begin{figure*}[t]
    \centering
    \includegraphics[width=\textwidth]{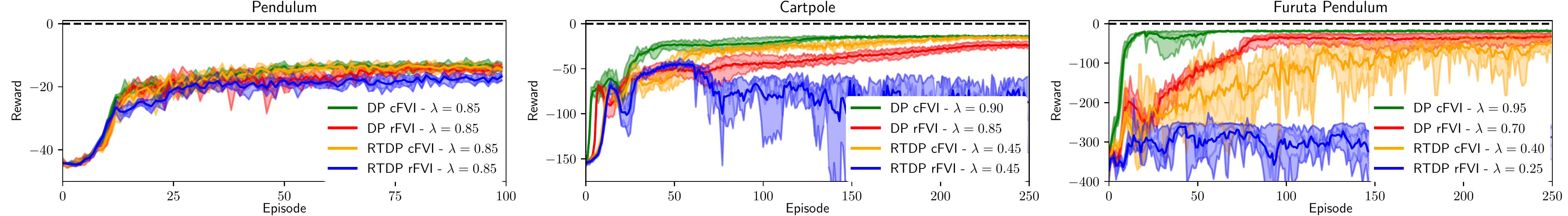}
    \vspace{-2.em}
    \caption{The learning curves for DP rFVI, DP cFVI, RTDP cFVI and RTDP rFVI averaged over $5$ seeds. The shaded area displays the \emph{min/max} range between seeds. DP rFVI learns slower compared to DP cFVI on the carpole and Furuta pendulum as the adversary prevents learning. RTDP rFVI does not learn the task as the adversary is too strong for the online variant of rFVI despite using the identical admissible set as the offline variant DP rFVI.} 
    \label{fig:learning_curves}\vspace{-0.8em}
\end{figure*} 

\medskip
\noindent
\textbf{Proof Sketch Theorem 3} \Eqref{eq:val_adv_update} can be written as
\begin{gather*}
V_{\text{tar}} = \max_{\vu} \: \min_{\vxi_{\theta}} \: r(\vx, \vu) + \gamma V\big(f(.)\big) 
\hspace{5pt} \text{s.t.} \hspace{5pt} \left(\vxi_{\theta} - \vmu_{\nu} \right)^{2} \leq  \vDelta_{\nu}^{2}
\end{gather*}
by replacing the admissible set $\Omega_{A}$ with an explicit constraint. In the following we abbreviate $\mB(\vx; \vtheta + \vxi_{\theta})$ as $\mB_{\xi}$ and $\va(\vx; \vtheta + \vxi_{\theta})$ as $\va_{\xi}$. 
Substituting the Taylor expansion for $V(\vx_{t+1})$, the dynamics models and reward yields 
\begin{align*}
\frac{V_\text{tar} - \gamma V}{\Delta t} &= q_c + \max_{\vu} \min_{\vxi} \left[\gamma \nabla_x V^{T} \left(\va_{\xi} + \mB_{\xi} \vu \right) + \gamma \mathcal{O}(.) - g_c \right]
\end{align*}
In the continuous-time limit the optimal action and disturbance is determined by 
\begin{align*}
\vu^{*}, \vxi^{*}_{\theta} = \max_{\vu} \min_{\vxi} \left[\nabla_x V^{T} \big(\va_{\xi} + \mB_{\xi} \vu \big) - g_c(\vu) \right].
\end{align*}
This nested max-min optimization can be solved by first solving the inner optimization w.r.t. to $\vu$ and substituting this solution into the outer maximization. The Lagrangian for the optimal model disturbance is described by
\begin{align*}
\vxi^{*} = \argmin_{\vxi} \: \nabla_x V^{T} \left(\va_{\xi} + \mB_{\xi} \vu \right) + \frac{1}{2} \vlambda^{T} \left((\vxi_{\theta} - \vmu_{\nu})^{2} - \vDelta_{\nu}^{2}\right).
\end{align*}
Using the KKT conditions this optimization can be solved. The stationarity condition yields
\begin{gather*}
\vz_{\theta} + \vlambda^{T} (\vxi_{\theta} - \vmu_{\nu}) \coloneqq 0  \hspace{15pt} \Rightarrow \hspace{15pt} \vxi^{*}_{\theta} = - \vz_{\theta} \oslash \vlambda + \vmu_{\nu} 
\end{gather*}
%
with the elementwise division $\oslash$. Using the primal feasibility and the complementary slackness, the optimal $\vlambda^{*}$ can be computed.
The resulting optimal model disturbance is described by 
\begin{align*}
\vxi^{*}_{\theta}(\vu) = 
-\vDelta_{\nu} \sign\big( \vz_{\theta}(\vu) \big) + \vmu_{\nu}
\end{align*}
as $\vz_{\theta} \oslash \| \vz_{\theta}\|_1 = \sign(\vz_{\theta})$. The action can be computed by 
\begin{align*}
\vu^{*} = \argmax_{\vu} \nabla_x V^{T} \left[\va(\vxi^{*}_{\theta}(\vu)) + \mB\left(\vxi^{*}_{\theta}(\vu)\right) \vu \right] - g_c(\vu).
\end{align*}
Due to the envelope theorem~\cite{carter2001foundations}, the extrema is described by
\begin{align*}
\mB(\vx; \vtheta + \vxi^{*}_{\theta}(\vu))^{T} \nabla_x V - g_{c}(\vu) \coloneqq 0.
\end{align*}
This expression cannot be solved without approximation as $\mB$ does not necessarily be invertible w.r.t. $\vtheta$. Approximating $\mB(\vx; \vtheta + \vxi^{*}(\vu)) \approx \mB(\vx; \vtheta)$, lets one solve for $\vu$. In this case the optimal action $\vu^{*}$ is described by  
$\vu^{*} {=} \nabla \tilde{g}(\mB(\vx; \vtheta)^{T} \nabla_x V)$. This approximation implies that neither agent or the adversary can react to the action of the other and must choose simultaneously. This assumption is common in prior works~\cite{bansal2017hamilton}. 
\hspace{\fill}\qed

\begin{figure*}[t]
    \centering
    \includegraphics[width=\textwidth]{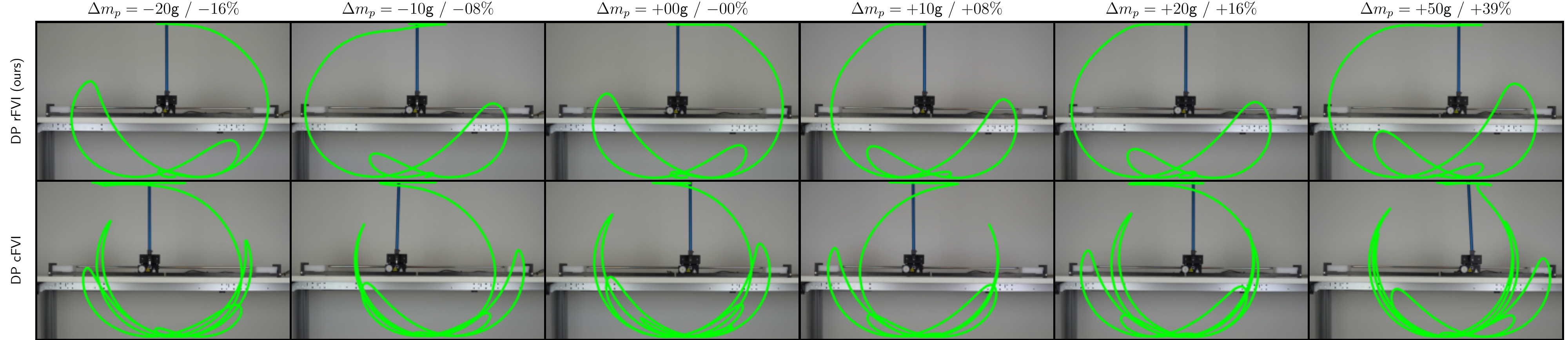}
    \vspace{-2.em}
    \caption{The tracked trajectories of DP rFVI and DP cFVI on the physical cartpole with varied pendulum masses. DP rFVI is capable to perform the swing-up for the varying pendulum mass. The qualitative performance does not change if the weight is added or reduced. DP cFVI can swing up and balance all varied pendulums but it requires more pre-swings for all configurations. During balancing the cart is not centered and the cart oscillates around the center for DP cFVI. The pendulum must significantly deviate from the target position before the DP cFVI policy breaks the stiction of the linear actuator. In contrast, the higher actions of DP rFVI break the stiction and balance the pendulum with the cart centered.}
    \label{fig:img_cartpole}\vspace{-0.8em}
\end{figure*} 

\subsection{Algorithm}
\noindent
Using the theoretical insights from the previous section, robust fitted value iteration can be derived. Instead of computing the value function target using only the optimal action as in cFVI, rFVI includes the four adversaries to learn a optimal policy that is robust to changes in the dynamics. Therefore, the value function target is computed using 
\begin{align}
V^{k+1}_{\text{tar}} &= \: r\left(\vx, \: \vu^{k} \right) + \gamma V^{k}\left(f\left(\vx, \: \vu^{k}, \: \vxi^{k}_{x}, \: \vxi^{k}_{u}, \: \vxi^{k}_{o}, \: \vxi^{k}_{\theta}\right)\right)
\end{align}
where actions $\vu^{k}$ and disturbances $\vxi_{i}$ are determined according to Table \ref{table:adversarial_disturbances}. Iterating between computing the target and fitting the value function network (\Eqref{eq:val_fitting}) enables the learning of the robust optimal value function and robust optimal policy. 

\medskip
\noindent
\textbf{$\mathbf{N}$-Step Value Function Target}
The learning can be accelerated by using the exponentially weighted $n$-step value target instead of the $1$-step target, as shown by the classical eligibility traces \cite{sutton1998introduction}, generalized advantage estimation \cite{schulman2015trust, schulman2017proximal} or model based value-expansion~\cite{feinberg2018model}. In the continuous limit this target is described by 
\begin{align*}
V_{\text{tar}}(\vx_0) &= \int_{0}^{T} \beta \: \exp(-\beta t) \: R_t \: dt + \exp(-\beta T) R_T,  \\ 
R_t &= \int_{0}^{t} \exp(-\rho \tau) \: r_c(\vx_{\tau}, \vu_{\tau}) d\tau + \exp(-\rho t) V^{k}(\vx_t) \label{eq:R_t}, \\
\vx_t &= \int_{0}^{t} f_{c}\left(\vx, \: \vu^{k}, \: \vxi^{k}_{x}, \: \vxi^{k}_{u}, \: \vxi^{k}_{o}, \: \vxi^{k}_{\theta}\right) \:\:d\tau + \vx_0,
\end{align*}
where $\beta$ is the exponential decay factor. In practice we treat $\beta$ as the hyperparameter and select $T$ such that the weight of the $R_T$ is $\exp\left(-\beta T\right) \coloneqq 10^{-4}$.

\medskip \noindent
\textbf{Admissible Set}
For the state, action and observation adversary the signal energy is bounded. We limit the energy of $\vxi_{x}$, $\vxi_{u}$ and $\vxi_{o}$ as the non-adversarial disturbances are commonly modeled as multivariate Gaussian distribution. Therefore, the average energy is determined by the noise covariance matrix. For the model parameters $\vtheta$ a common practice is to assume that the approximate model parameters have an model error of up to~$\pm15\%$ \cite{muratore2018domain, muratore2021data}. Hence, we bound the amplitude of each component. To not overfit to the deterministic worst case system of $V^k$ and enable the discovery of good actions, the amplitude of the adversarial actions of $\xi_x$, $\xi_u$, $\xi_o$ is modulated using a Wiener process. This random process allows a continuous-time formulation that is agnostic to the sampling frequency.

\medskip \noindent
\textbf{Offline and Online rFVI}
The proposed approach is off-policy as the samples in the replay memory do not need to originate from the current policy $\pi_k$. Therefore, the dataset can either consist of a fixed dataset or be updated within each iteration. In the offline dynamic programming case, the dataset contains samples from the compact state domain~$\mathcal{X}$. We refer to the offline variant as DP rFVI. In the online case, the replay memory is updated with samples generated by the current policy $\pi_k$. Every iteration the states of the previous $n$-rollouts are added to the data and replace the oldest samples. 
This online update of state distribution performs real-time dynamic programming (RTDP)~\cite{barto1995learning}. We refer to the online variant as RTDP rFVI. The pseudo code of DP cFVI and RTDP rFVI is summarized in Algorithm~\ref{alg:rFVI}.

\section{Experiments} \label{sec:experiments}
\noindent
In the following non-linear control experiments we want to answer the following questions:

\smallskip \noindent
\textbf{Q1:} Does rFVI learn a robust policy that can be successfully transferred to the physical systems with different model parameters? 

\noindent
\textbf{Q2:} How does the policy obtained by rFVI differ qualitatively compared to cFVI and the deep RL baselines? 

\begin{figure}[t]
    \centering
    \includegraphics[width=\columnwidth]{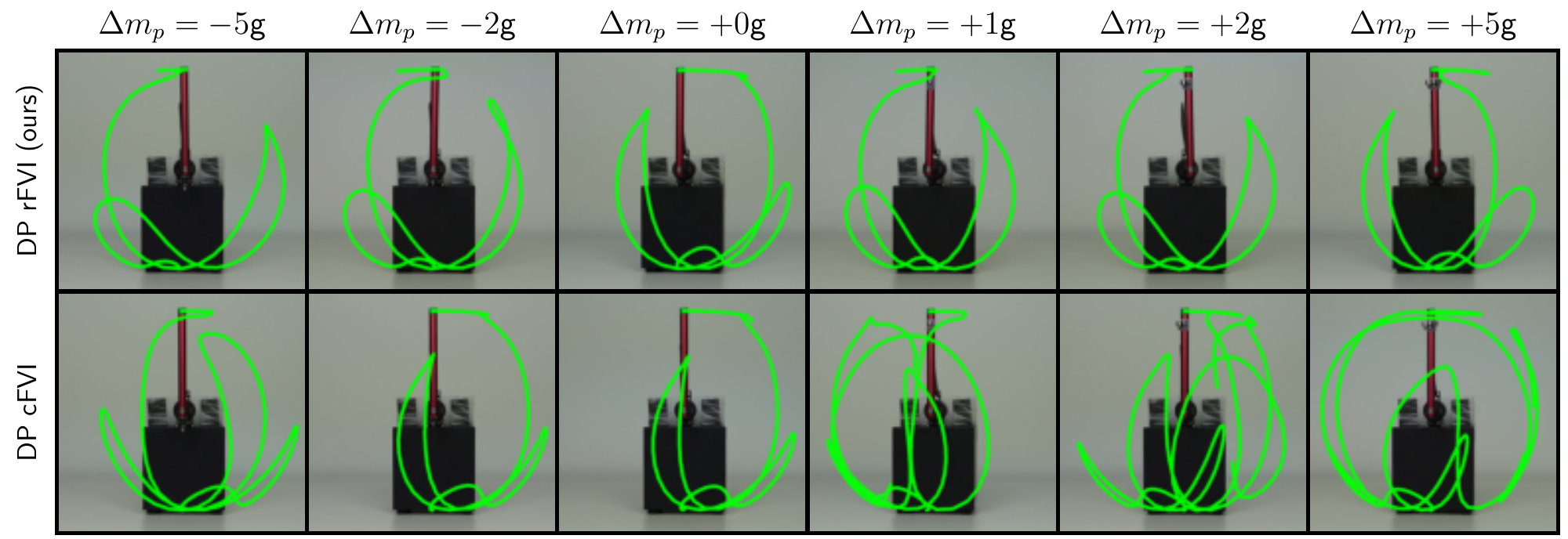}
    \vspace{-2.em}
    \caption{The tracked trajectories for DP rFVI and DP cFVI on the Furuta pendulum for different pendulum weights. The trajectories of rFVI do not significantly change when the pendulum mass altered. For DP cFVI the trajectories start to change when an additional weight is added. For these system dynamics, DP cFVI requires some failed swing-ups until the policy can balance the pendulum.}
    \label{fig:img_furuta}\vspace{-0.8em}
\end{figure}

\smallskip\noindent
To answer these questions, we apply the algorithms to perform the swing-up task of the under-actuated cartpole (Fig. \ref{fig:img_cartpole}) and Furuta pendulum (Fig. \ref{fig:img_furuta}). Both systems are standard environments for benchmarking non-linear control policies. We focus only on these two systems to perform extensive robustness experiments on the actual physical systems. To test the robustness with respect to model parameters, we attach small weights to the passive pendulum.  

\begin{figure*}[t]
    \centering
    \includegraphics[width=\textwidth]{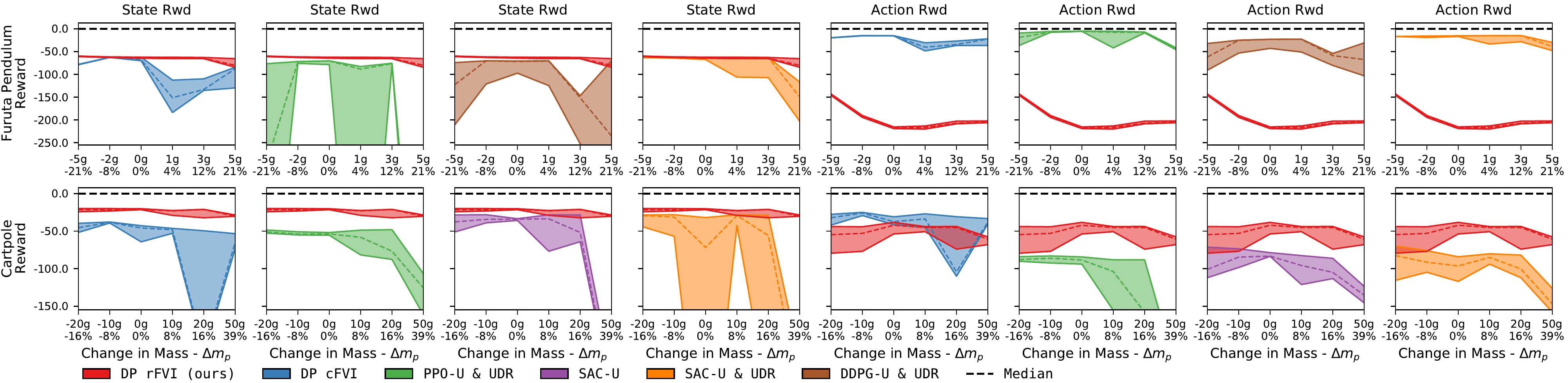}
    \vspace{-2.em}
    \caption{The $25$th, $50$th and $75$th reward percentile for the physical Furuta pendulum and cartpole with varied pendulum weights. DP rFVI achieves higher state reward for real-world systems compared to the baselines. For the different weights the reward remains nearly constant. For the Furuta pendulum the action cost is significantly higher compared to the baselines as the DP rFVI causes a chattering during balancing due to the high actions and minor time delays in the control loop. If only the swing-up phase is considered the rewards are comparable.}
    \label{fig:furuta_trajectories}\vspace{-0.8em}
\end{figure*} 

\subsection{Experimental Setup}
\noindent 
\textbf{Systems} The physical cartpole and Furuta pendulum are manufactured by Quanser \cite{quanser} and voltage controlled. For the approximate simulation model we use the rigid-body dynamics model with the parameters supplied by the manufacturer. If we add negative weights to the pendulum, we attach the weights to the opposite lever of the pendulum. This moves the center of mass of the pendulum closer to the rotary axis. Therefore, this shift reduces the downward force and is equivalent to a lower pendulum mass.

\medskip
\noindent
\textbf{Baselines} The performance is compared to the actor-critic deep RL methods: DDPG \cite{lillicrap2015continuous}, SAC \cite{haarnoja2018soft} and PPO \cite{schulman2017proximal}. The robustness evaluation is only performed for the best performing baselines on the nominal physical system. The performance of all baselines on the nominal system is summarized in Table \Romannum{4} (Appendix). 
The initial state distribution is abbreviated by \{SAC, PPO, DDPG\}-U for a uniform distribution of the pendulum angle and \{SAC, PPO, DDPG\}-N for a Gaussian distribution. The baselines with Gaussian initial state distribution did not achieve robust performance on the nominal system. If the baseline uses uniform domain randomization the acronym is appended with UDR.  

\medskip
\noindent
\textbf{Evaluation}
To evaluate rFVI and the baselines we separately compare the state and action reward as these algorithms optimize a different objectives. Hence, these algorithms trade-off state and action associated rewards differently. It is expected that the worst-case optimization uses higher actions to prevent deviation from the optimal trajectory. On the physical system, the performance is evaluated using the $25$th, $50$th and $75$th~reward percentile as the reward distribution is multi-modal. 

\subsection{Experimental Results}
\noindent
The learning curves averaged over 5 seeds of DP rFVI and RTDP rFVI are shown in Figure \ref{fig:learning_curves}. The results in simulation are summarized in Table \ref{table:simulation_results}. DP rFVI learns a policy that obtains slightly lower reward compared to cFVI and the deep RL baselines. This lower reward is expected as the worst-case optimization yields conservative policies \cite{xu2012robustness}. DP rFVI exhibits low variance between seeds but learns slower than DP cFVI. This slower learning is caused by the adversary which counteracts the learning progress. RTDP rFVI does not learn to successfully swing-up the Furuta pendulum and cartpole. Despite using the same admissible set for both variants, the adversary is too powerful for RTDP rFVI and prevents learning. In this case policy does not discover the positive reward at the top as the adversary prevents balancing. Therefore, the policy is too pessimistic and converges to a bad local optima. The ablation study in the appendix shows, that RTDP rFVI learns a successful policy if the admissible sets are reduced. To overcome this problem one would need to bias the exploration to be optimistic. 

\begin{table}[t]
\tiny
\centering
\renewcommand{\arraystretch}{1.25}
\caption{\footnotesize
The average rewards in simulation.\vspace{-10pt}
}
\setlength{\tabcolsep}{2.5pt}
\begin{tabular*}{\columnwidth}{l |
c c |
c c |
c c
}
\toprule
& \multicolumn{2}{c|}{Sim Pendulum}  
& \multicolumn{2}{c|}{Sim Cartpole}
& \multicolumn{2}{c}{Sim Furuta Pendulum}  
\\
Algorithm 
& State Rwd & Action Rwd
& State Rwd & Action Rwd
& State Rwd & Action Rwd
\\
\cmidrule(lr){1-1} \cmidrule(lr){2-3} \cmidrule(lr){4-5} \cmidrule(lr){6-7}
DP rFVI (ours)
& $-24.5 \pm 00.1$ & $\mathbf{-08.3 \pm 00.4}$
& $\mathbf{-15.5 \pm 05.4}$ & $-11.6 \pm 02.0$
& $-37.0 \pm 13.2$ & $-04.8 \pm 02.7$
\\
DP cFVI
& $-22.3 \pm 03.0$ & $\mathbf{-08.3 \pm 03.2}$
& $\mathbf{-14.4 \pm 02.8}$ & $-09.9 \pm 02.2$
& $\mathbf{-22.3 \pm 02.5}$ & $-05.9 \pm 01.3$
\\
SAC-U
& $\mathbf{-21.1 \pm 05.2}$ & $-09.4 \pm 04.7$
& $\mathbf{-13.9 \pm 02.4}$ & $-10.4 \pm 02.3$
& $\mathbf{-22.6 \pm 02.9}$ & $-05.9 \pm 01.5$
\\
SAC-U \& UDR
& $-22.6 \pm 02.6$ & $-08.9 \pm 01.9$
& $\mathbf{-13.9 \pm 02.6}$ & $-10.3 \pm 02.5$
& $\mathbf{-22.1 \pm 02.6}$ & $-06.1 \pm 01.4$
\\
DDPG-U
& $\mathbf{-20.9 \pm 01.2}$ & $-10.6 \pm 01.3$
& $-16.4 \pm 04.7$ & $-11.8 \pm 04.9$
& $\mathbf{-24.6 \pm 03.6}$ & $-05.5 \pm 01.9$
\\
DDPG-U \& UDR
& $\mathbf{-21.0 \pm 04.8}$ & $-11.5 \pm 01.6$
& $\mathbf{-14.5 \pm 03.5}$ & $-12.8 \pm 03.5$
& $-26.1 \pm 02.4$ & $-06.2 \pm 01.9$
\\
PPO-U
& $-24.5 \pm 04.4$ & $-09.0 \pm 02.4$
& $-82.2 \pm 83.1$ & $\mathbf{-04.9 \pm 01.1}$
& $-34.9 \pm 12.6$ & $\mathbf{-03.4 \pm 01.1}$
\\
PPO-U \& UDR
& $-24.5 \pm 00.2$ & $-11.1 \pm 03.0$
& $-55.9 \pm 23.1$ & $-09.8 \pm 05.2$
& $-42.9 \pm 04.6$ & $-06.1 \pm 02.4$
\\
\bottomrule
\end{tabular*}
\vspace{-2.5em}
\label{table:simulation_results}
\end{table}

\medskip \noindent
The performance on the physical systems is summarized in Figure \ref{fig:img_cartpole}, \ref{fig:img_furuta} and \ref{fig:furuta_trajectories}.\footnote{Videos of the experiments at \textcolor{blue}{\url{https://sites.google.com/view/rfvi}}.} Across the different parameters of both systems, rFVI achieves the highest state reward compared to cFVI and the deep RL baselines. The best performing trajectories between different configurations are nearly identical. Only for the cartpole the failure rates slightly increases when positive weights are added. In this case the pendulum is swung-up but cannot be stabilized, due to the backslash of the linear actuator. For cFVI and the deep RL baselines even the best trajectories deteriorate when weights are added to the pendulum. This is especially notable in Figure \ref{fig:furuta_trajectories}, where the deep RL baselines start to explore the complete state space when additional weights are added. The high state rewards of rFVI are obtained at the expense of higher action cost, which can be higher compared to some baselines on the physical system. To summarize rFVI obtains a robust policy that performs the swing-up with low state reward and more consistency than the baselines but uses higher actions.

\medskip \noindent
Compared to the policies obtained by DP cFVI and the deep RL baselines, DP rFVI policy exerts higher actions. Therefore, DP rFVI achieves the robustness compared to the baselines by utilizing a stiffer feedback policy approaching bang-bang control. The higher actions are only observed on the physical system where the deviation from the optimal trajectory is inevitable. In simulation the action cost are comparable to the baselines. Therefore, the high actions originate from the feedback-term of the non-linear policy that tries to compensate the tracking error. The stiffer feedback policy is expected as traditional robust control approaches yield high feedback gains \cite{aastrom1987comparison}. The stiffness of the rFVI policy is clearly visible in Figure \ref{fig:value_fun_pendulum}. On the ridge leading up to the balancing point, the policy directly applies maximum action, if one deviates from the center of the ridge. In contrast, DP cFVI has a gradient that slightly increases the action when one deviates from the optimal trajectory. 

\begin{figure*}[t]
    \centering
    \includegraphics[width=\textwidth]{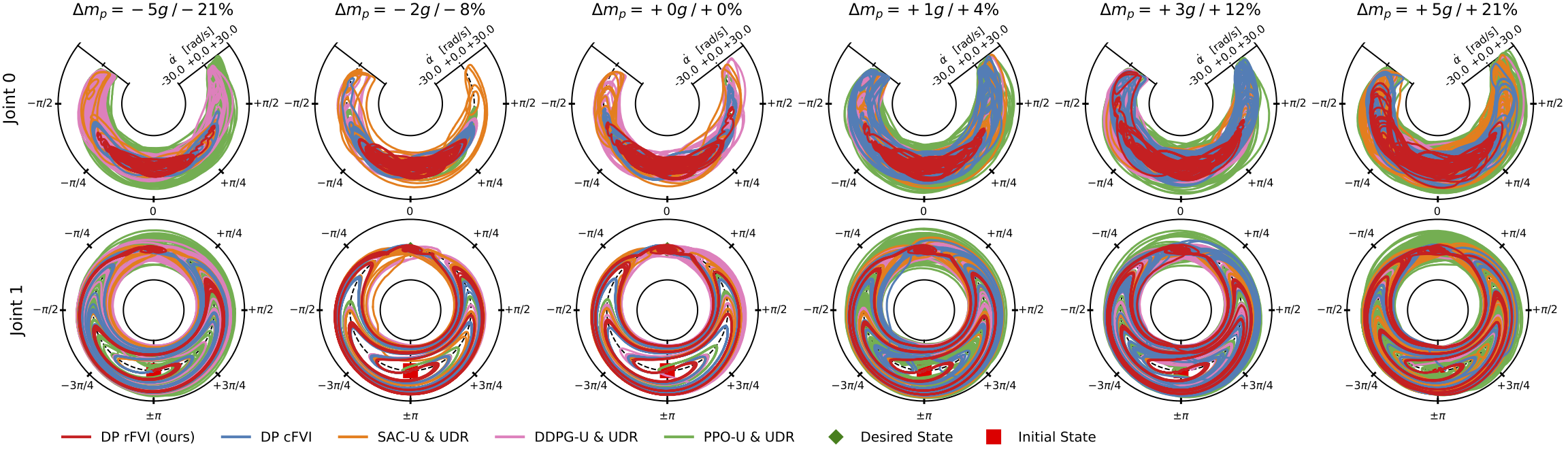}
    \vspace{-2.em}
    \caption{The roll-outs of DP rFVI, DP cFVI and the deep RL baselines with domain randomization the physical Furuta pendulum. The different columns correspond to different pendulum masses. The deviation from the dashed center line corresponds to the joint velocity. DP rFVI achieves a consistent swing-up for the different pendulum masses. In contrast to DP rFVI, the baselines start to deviate strongly from trajectories on the nominal system. When weights are added the baselines start to cover the complete state-space. A figure displaying the roll-outs per algorithm is provided in the appendix.}
    \label{fig:furuta_trajectories}\vspace{-0.8em}
\end{figure*} 

\medskip \noindent
For the cartpole the higher actions achieve much better performance. DP rFVI achieves stabilization of the cart in the center of the track (Figure \ref{fig:img_cartpole}). The higher actions break the stiction and overcome the backslash of the geared cogwheel actuator during the balancing. For DP cFVI and the baselines, the cart oscillates around the center. The pendulum angle has to deviate significantly until the actions become large and break the stiction. For the Furuta pendulum the high actions are more robust during the swing-up phase. However, during the balancing the lower link starts to chatter due to the high-frequency switching between high actions. This switching is caused by minor delays in the control loop and the very low friction of the pendulum. About $90\%$ of the action cost of DP rFVI for the Furuta pendulum is incurred during the stabilization. DP~cFVI incurs only $10\%$ of the action cost during stabilization. If one would only consider the swing-up phase, the reward of DP rFVI is higher compared to the baselines. To summarize, high-stiffness feedback policies are robust to changes in dynamics. However, this robustness can also make the resulting policies more sensitive to other sources error not included in the specification, e.g., control delays.

\medskip \noindent 
Besides the performance evaluation of DP rFVI, the experiments show that DP~cFVI achieves comparable performance than the deep RL baselines with domain randomization. Despite overfitting to a deterministic approximate model, DP~cFVI is able to be robust against some variations in model parameters. This result suggests that for these two physical systems, preventing the distribution shift by solving for the policy on the compact state domain obtains a policy with comparable robustness as uniform domain randomization. Furthermore, the deep RL performance increases with larger state distribution and using the maximum-entropy formulation on the physical system. Therefore, the experiments suggest that the state-distribution of the policy affects the policy robustness. This correlation should be investigated in-depth in future work.

\section{Related Work}
\noindent
\textbf{Robust Policies} Learning robust policies has been approached by (1) changing the optimization objective that balances risk and reward \cite{borkar2001sensitivity, chow2015risk,bharadhwaj2021csc}, (2) introducing a adversary to optimize the worst-case performance~\cite{tamar2013scaling, isaacs1999differential, bansal2017hamilton, pinto2017robust,harrison2017adapt} and (3) randomizing the dynamics model to be robust to the model parameters~\cite{andrychowicz2020learning, muratore2021data, muratore2021data, xie2021dynamics, chebotar2019closing, ramos2019bayessim}. In robotics, domain randomization is most widely used to achieve successful sim2real transfer. For example domain randomization was used for in-hand manipulation~\cite{andrychowicz2020learning}, ball-in-a-cup~\cite{muratore2021data}, locomotion~\cite{xie2021dynamics}, manipulation~\cite{chebotar2019closing, ramos2019bayessim}. 

\medskip \noindent
In this paper we focus on the adversarial formulation. This approach has been extensively used for continuous control tasks~\cite{morimoto2005robust, pinto2017robust, pinto2017supervision, mandlekar2017adversarially, tessler2019action}. For example, Pinto et. al. \cite{pinto2017robust, pinto2017supervision} used a separate agent as adversary controlling an additive control input. This adversary maximized the negative reward using a standard actor-critic learning algorithm. The agent and adversary do not share any information. Therefore, an additional optimization is required to optimize the adversary. A different approach by Mandlekar et. al. \cite{mandlekar2017adversarially} used auxiliary loss to maximize the policy actions. Both approaches are model-free. In contrast to these approaches, our approach derives the adversarial perturbations using analytic expressions derived directly from the Hamilton-Jacobi-Isaacs (HJI) equation. Therefore, our approach shares knowledge between the actor and adversary due to a shared value function and requires no additional optimization. However, our approach requires knowledge of the model to compute the actions and perturbations analytically. The approach is similar to Morimoto and Doya \cite{morimoto2005robust}. In contrast to this work, we extend the analytic solutions to state, action, observation and  model disturbances, do not require a control-affine disturbance model, and use the constrained formulation rather than the penalized formulation.

\medskip \noindent
\textbf{Continuous-Time Reinforcement Learning}
Various approaches have been proposed to solve the Hamilton-Jacobi-Bellman (HJB) differential equation with the machine learning toolset. These methods can be divided into trajectory and state-space based methods. 
Trajectory based methods solve the stochastic HJB along a trajectory using path integral control \cite{kappen2005linear, todorov2007linearly, theodorou2010reinforcement} or forward-backward stochastic differential equations \cite{pereira2019learning, pereira2020safe}.
State-space based methods solve the HJB globally to obtain a optimal non-linear controller applicable on the complete state domain. Classical approaches discretize the problem and solve the HJB or HJI using a PDE solver~\cite{bansal2017hamilton}. To overcome the curse of dimensionality of the grid based methods, machine learning methods have proposed to use various function approximators ranging from polynomial functions \citep{yang2014reinforcement, liu2014neural}, kernels \cite{hennig2011optimal} to deep networks \cite{doya2000reinforcement, tassa2007least, lutter2019hjb, kim2020hamilton}. 
In this paper, we utilize the state-space based approach solve the HJI via dynamic programming. The value function is approximated using a deep network and optimal value function is solved via value iteration. 

\section{Conclusion and Future Work}
\noindent
We proposed robust fitted value iteration (rFVI). This algorithm solves the adversarial continuous-time reinforcement learning problem for continuous states, action and adversary via value iteration. To enable the efficient usage of value iteration, we presented  analytic expressions for the adversarial disturbances for the state, action, observation and model adversary. Therefore, our derivations extend existing analytic expressions for continuous time RL from  literature \cite{lyshevski1998optimal, doya2000reinforcement, morimoto2005robust, lutter2019hjb}. The non-linear control experiments using the physical cartpole and Furuta pendulum showed that rFVI is robust to variations in model parameters and obtains higher state-rewards compared to the deep RL baselines with uniform domain randomization. The robustness of rFVI is achieved by utilizing a stiffer feedback policy that exerts higher actions compared to the baselines. 

\medskip \noindent
In future work we plan to learn the admissible sets from data from the physical system. In domain randomization, the automatic tuning of the parameter distributions has been very successful \cite{ramos2019bayessim, muratore2021data}. However, these approaches are not directly transferable as one would also need to estimate the admissible set of the action, state and observation and not only the system parameter distribution as in domain randomization.

{\small
\section*{Acknowledgements}
\noindent M. Lutter was an intern at Nvidia during this project. 
A. Garg was partially supported by CIFAR AI Chair. 
We also want to thank Fabio Muratore, Joe Watson and the RSS reviewers for their feedback. Furthermore, we want to thank the open-source projects SimuRLacra~\cite{simurlacra}, MushroomRL~\cite{deramo2020mushroomrl}, NumPy~\cite{numpy} and PyTorch~\cite{pytorch}.
}

\bibliographystyle{ieeetr}
\bibliography{refs}

\textcolor{white}{Next Page}
\clearpage

\section*{Appendix}

\subsection{State Disturbance Proof}
\begin{theorem*}
For the adversarial state disturbance (\Eqref{eq:dyn_adv_state}) with bounded in signal energy (\Eqref{eq:signal_energy}), the optimal continuous time policy $\pi^{k}$ and state disturbance $\vxi^{k}_{x}$ is described by
\begin{align*}
\pi^{k}(\vx) &= \nabla \tilde{g} \left(\mB(\vx)^{T} \nabla_x V^{k}\right) & \vxi^{k} &= - \alpha \frac{\nabla_x V^{k}}{\| \nabla_x V^{k} \|_2}.
\end{align*}
\end{theorem*}

\noindent
\emph{Proof.} \Eqref{eq:val_adv_update} can be formulated with the explicit constraint, i.e., 
\begin{gather*}
V_{\text{tar}} = \max_{\vu} \: \min_{\vxi_x} \: r(\vx_{t}, \vu) + \gamma V\big(f(\vx_{t}, \vu, \: \vxi_x)\big) \\
\text{with} \hspace{5pt} \vxi_{x}^{T} \vxi_{x} - \alpha^{2} \leq 0. 
\end{gather*}
Substituting the Taylor expansion, the dynamics model and the reward yields
\begin{align*}
V_{\text{tar}} &= \max_{\vu} \: \min_{\vxi_x} \: r + \gamma V + \gamma \nabla_x V^{T} f_c \Delta t + \gamma \mathcal{O} \Delta t \\
\frac{V_\text{tar} - \gamma V}{\Delta t} &= q_c + \max_{\vu} \left[\nabla_x V^{T} \left(\va + \mB \vu \right) + \gamma \mathcal{O} - g_c \right] \\
&\hspace{99pt} + \min_{\vxi} \left[\nabla_x V^{T} \vxi_x \right]
\end{align*}
with the higher order terms $\mathcal{O}(\vx, \vu, \Delta t)$. In the continuous time limit, the higher-order terms disappear, i.e., $\lim_{t\rightarrow 0} \mathcal{O} = 0$. Therefore, the optimal action is described by 
\begin{gather*}
\vu_{t} = \argmax_{\vu} \:  \nabla_x V^{T} \mB(\vx_t) \: \vu - g_{c}(\vu) \\
\Rightarrow \hspace{5pt}
\vu_{t} = \nabla \tilde{g}_c\left(\mB(\vx_t) \nabla_x V\right).
\end{gather*}
The optimal state disturbance is described by
\begin{align*}
\vxi^{*}_x = \argmin_{\vxi_x} \: \nabla_x V^{T} \vxi_x 
\hspace{10pt} \text{with} \hspace{10pt} 
\vxi_{x}^{T} \vxi_{x} - \alpha^{2} \leq 0.
\end{align*}
This constrained optimization can be solved using the Karush-Kuhn-Tucker (KKT) conditions, i.e., 
\begin{align*}
\nabla_x V + 2 \lambda \: \vxi_x = 0 \hspace{10pt} \Rightarrow \hspace{10pt} \vxi^{*}_{x} = -\frac{1}{2 \lambda} \nabla_x V
\end{align*}
with the Lagrangian multiplier $\lambda \geq 0$. From primal feasibility and the complementary slackness condition of the KKT conditions follows that
\begin{align*}
\frac{1}{4\lambda^{2}} \nabla_x V^{T} \nabla_x V - \alpha^{2} &\leq 0 \hspace{0pt}  \\
& \hspace{45pt} \Rightarrow \hspace{2pt} \lambda \geq \frac{1}{2 \alpha} \sqrt{\nabla_x V^{T} \nabla_x V}\\
\lambda \left(\frac{1}{4 \lambda^{2}} \nabla_x V^{T} \nabla_x V - \alpha^{2}\right) &= 0 \hspace{0pt}\\ 
&\hspace{5pt} \Rightarrow \hspace{2pt} \lambda_0 = 0, \hspace{5pt} \lambda_1 = \frac{1}{2 \alpha} \sqrt{\nabla_x V^{T} \nabla_x V}.
\end{align*}
Therefore, the optimal adversarial state perturbation is described by
\begin{align*}
\vxi^{k} = - \alpha \frac{V^{k}_{x}}{\| V^{k}_{x} \|_2}.
\end{align*}
This solution is intuitive as the adversary wants to minimize the reward and this disturbance performs steepest descent using the largest step-size. Therefore, the perturbation is always on the constraint. \hspace{\fill}\mbox{\qed}

\subsection{Action Disturbance Proof}
\begin{theorem*}
For the adversarial action disturbance (\Eqref{eq:dyn_adv_state}) with bounded in signal energy (\Eqref{eq:signal_energy}), the optimal continuous time policy $\pi^{k}$ and action disturbance $\vxi^{k}_{u}$ is described by
\begin{align*}
\pi^{k}(\vx) &= \nabla \tilde{g} \left(\mB(\vx)^{T} \nabla_x V^{k}\right) & \vxi^{k}_{u} &= - \alpha \frac{\mB(\vx)^{T} \nabla_x V^{k}}{\| \mB(\vx)^{T} \nabla_x V^{k} \|_2}.
\end{align*}
\end{theorem*}

\noindent
\emph{Proof.} \Eqref{eq:val_adv_update} can be formulated with the explicit constraint, i.e., 
\begin{align*}
V_{\text{tar}} = \max_{\vu} \: \min_{\vxi_u} \: r(\vx, \vu) + \gamma V\big(f(\vx, \vu, \: \vxi_u)\big) \hspace{5pt} \text{with} \hspace{5pt} \vxi_{u}^{T} \vxi_{u} \leq \alpha^{2}. 
\end{align*}
Substituting the Taylor expansion, the dynamics model and the reward yields
\begin{align*}
V_{\text{tar}} &= \max_{\vu} \: \min_{\vxi_u} \: r + \gamma V + \gamma \nabla_x V^{T} f_c \Delta t + \gamma \mathcal{O} \Delta t \\
\frac{V_\text{tar} - \gamma V}{\Delta t} &= q_c + \max_{\vu} \left[\nabla_x V^{T} \left(\va + \mB \vu \right) + \gamma \mathcal{O} - g_c \right] \\
&\hspace{88pt}+ \min_{\vxi} \left[\nabla_x V^{T} \mB \: \vxi_u \right]
\end{align*}
with the higher order terms $\mathcal{O}(\vx, \vu, \Delta t)$. In the continuous time limit, the higher-order terms disappear, i.e., $\lim_{t\rightarrow 0} \mathcal{O} = 0$. Therefore, the optimal action is described by 
\begin{gather*}
\vu_{t} = \argmax_{\vu} \:  \nabla_x V^{T} \mB(\vx) \: \vu - g_{c}(\vu)  \\
\Rightarrow \hspace{2pt}
\vu_{t} = \nabla \tilde{g}_c\left(\mB(\vx) \nabla_x V\right).
\end{gather*}
The optimal state disturbance is described by
\begin{align*}
\vxi^{*}_u = \argmin_{\vxi_u} \: \nabla_x V^{T} \mB(\vx) \: \vxi_u \hspace{10pt} \text{with} \hspace{10pt} \vxi_{u}^{T} \vxi_{u} \leq \alpha^{2}.
\end{align*}
This constrained optimization can be solved using the Karush-Kuhn-Tucker (KKT) conditions, i.e., 
\begin{align*}
\mB(\vx_t)^{T} \nabla_x V + 2 \lambda \: \vxi_u = 0 \hspace{10pt} \Rightarrow \hspace{10pt} \vxi^{*}_{u} = -\frac{1}{2 \lambda} \mB(\vx_t)^{T} \nabla_x V
\end{align*}
with the Lagrangian multiplier $\lambda \geq 0$. From primal feasibility and the complementary slackness condition of the KKT conditions follows that
\begin{align*}
\frac{1}{4 \lambda^{2}} \nabla_x V^{T} \mB \mB^T \nabla_x V - \alpha^{2} &\leq 0 \\
&\hspace{5pt} \Rightarrow \hspace{2pt} \lambda \geq \frac{1}{2 \alpha} \sqrt{\nabla_x V^{T} \mB \mB^{T} \nabla_x V}\\
\lambda \left(\frac{1}{4 \lambda^{2}} \nabla_x V^{T} \mB \mB^{T} \nabla_x V - \alpha^{2}\right) &= 0 \\
&\hspace{-35pt} \Rightarrow \hspace{2pt} \lambda_0 = 0, \hspace{5pt} \lambda_1 = \frac{1}{2\alpha} \sqrt{\nabla_x V^{T} \mB \mB^{T} \nabla_x V}.
\end{align*}
Therefore, the optimal adversarial state perturbation is described by
\begin{align*}
\vxi^{k}_{u} = - \alpha \frac{\mB(\vx_t)^{T} V^{k}_{x}}{\| \mB(\vx_t)^{T} V^{k}_{x} \|_2}.
\end{align*}
\hspace{\fill}\mbox{\qed}

\subsection{Model Disturbance Proof}
\begin{theorem*}
For the adversarial model disturbance (\Eqref{eq:dyn_adv_model}) with element-wise bounded amplitude (\Eqref{eq:amplitude}), smooth drift and control matrix (i.e., $\va, \mB \in C^{1}$) and $\mB(\theta + \vxi_{\theta}) \approx \mB(\theta)$, the optimal continuous time policy $\pi^{k}$ and model disturbance~$\vxi^{k}_{\theta}$ is described by
\begin{gather*}
\pi^{k}(\vx) = \nabla \tilde{g} \left(\mB(\vx)^{T} \nabla_x V^{k} \right) \hspace{20pt}
\vxi^{k}_{\theta} = -\vDelta_{\nu} \sign \left( \vz_{\theta }\right) + \vmu_{\nu} \\
\text{with} \hspace{5pt} \vz_{\theta} = \left(\frac{\partial \mB}{\partial \theta} \vu^{*} + \frac{\partial \va}{\partial \theta} \right)^T V^{*}_{x},
\end{gather*}
mean $\vmu_{\vnu} = \left( \vnu_{\text{max}} + \vnu_{\text{min}} \right) / 2$ and range $\vDelta_{\vnu} = \left( \vnu_{\text{max}} - \vnu_{\text{min}} \right) / 2$.
\end{theorem*}

\medskip
\noindent
\emph{Proof.} \Eqref{eq:val_adv_update} can be written with the explicit constraint instead of the infimum with the admissible set $\Omega_{A}$, i.e., 
\begin{gather*}
V_{\text{tar}} = \max_{\vu} \: \min_{\vxi_{\theta}} \: r(\vx, \vu) + \gamma V\big(f(\vx, \vu, \: \vxi_{\theta})\big) \\ 
\text{with} \hspace{5pt} \frac{1}{2}\left((\vxi_{\theta} - \vmu_{\nu})^{2} - \vDelta_{\nu}^{2}\right) \leq \mathbf{0}.
\end{gather*}
Substituting the Taylor expansion for $V(\vx_{t+1})$, the dynamics models and reward as well as abbreviating $\mB(\vx; \theta + \vxi_{\theta})$ as $\mB_{\xi}$ and $\va(\vx; \theta + \vxi_{\theta})$ as $\va_{\xi}$ yields
\begin{align*}
V_{\text{tar}} &= \max_{\vu} \: \min_{\vxi_{\theta}} \: r + \gamma V + \gamma \nabla_x V^{T} f_c \Delta t + \gamma \mathcal{O}(.) \Delta t \\
\frac{V_\text{tar} - \gamma V}{\Delta t} &= q_c + \max_{\vu} \min_{\vxi} \left[\gamma \nabla_x V^{T} \left(\va_{\xi} + \mB_{\xi} \vu \right) + \gamma \mathcal{O}(.) - g_c \right]
\end{align*}
In the continuous time limit the optimal action and disturbance is determined by 
\begin{align*}
\vu^{*}, \vxi^{*}_{\theta} = \max_{\vu} \min_{\vxi} \left[\nabla_x V^{T} \left(\va_{\xi} + \mB_{\xi} \vu \right) - g_c(\vu) \right].
\end{align*}
This nested max-min optimization can be solved by first solving the inner optimization w.r.t. to $\vu$ and substituting this solution into the outer maximization. The Lagrangian for the optimal model disturbance is described by
\begin{align*}
\vxi^{*} = \argmin_{\vxi} \: \nabla_x V^{T} \left(\va_{\xi} + \mB_{\xi} \vu \right) + \frac{1}{2} \vlambda^{T} \left((\vxi_{\theta} - \vmu_{\nu})^{2} - \vDelta_{\nu}^{2}\right)
\end{align*}
Using the KKT conditions this optimization can be solved. The stationarity condition yields
\begin{gather*}
\vz_{\theta} + \vlambda^{T} (\vxi_{\theta} - \vmu_{\nu}) = 0  \hspace{15pt} \Rightarrow \hspace{15pt} \vxi^{*}_{\theta} = - \vz_{\theta} \oslash \vlambda + \vmu_{\nu} \\
\text{with} \hspace{5pt} \vz_{\theta} =  \left[ \frac{\partial \va}{\partial \theta} + \frac{\partial \mB}{\partial \theta} \vu \right]^{T} \nabla_x V
\end{gather*}
%
and the elementwise division $\oslash$. The primal feasibility and the complementary slackness yields
\begin{align*}
\frac{1}{2} \left(-\vz_{\theta}^{2} \oslash \vlambda^{2} - \vDelta_{\nu}^{2} \right) \leq 0 \hspace{5pt} &\Rightarrow \hspace{5pt}  
\vlambda  \geq \| \vz_{\theta} \|_{1} \oslash \vDelta_{\nu} \\
\frac{1}{2} \vlambda^{T} \left(-\vz_{\theta}^{2} \oslash \vlambda^{2} - \vDelta_{\nu}^{2} \right) = 0 \hspace{5pt} &\Rightarrow \hspace{5pt} \vlambda_{0} = \mathbf{0}, \hspace{5pt} \vlambda_1 = \| \vz_{\theta} \|_{1} \oslash \vDelta_{\nu}.
\end{align*}
Therefore, the optimal model disturbance is described by 
\begin{align*}
\vxi^{*}_{\theta}(\vu) = 
-\vDelta_{\nu} \sign\big( \vz_{\theta}(\vu) \big) + \vmu_{\nu}
\end{align*}
as $\vz_{\theta} \oslash \| \vz_{\theta}\|_1 = \sign(\vz_{\theta})$. Then the optimal action can be computed by 
\begin{align*}
\vu^{*} = \argmax_{\vu} \nabla_x V^{T} \left[\va(\vxi^{*}_{\theta}(\vu)) + \mB\left(\vxi^{*}_{\theta}(\vu)\right) \vu \right] - g_c(\vu).
\end{align*}
Due to the envelope theorem, the extrema is described by
\begin{align*}
\mB(\vx; \theta + \vxi^{*}(\vu))^{T} \nabla_x V - g_{c}(\vu) = 0.
\end{align*}
This expression cannot be solved without approximation as $\mB$ does not necessarily be invertible w.r.t. $\theta$. Approximating $\mB(\vx; \theta + \vxi^{*}(\vu)) \approx \mB(\vx; \theta)$, lets one solve for $\vu$. In this case the optimal action $\vu^{*}$ is described by  
$\vu^{*} = \nabla \tilde{g}(\mB(\vx; \theta)^{T} \nabla_x V)$. This approximation is feasible for two reasons. First of all, if the adversary can significantly alter the dynamics in each step, the system would not be controllable and the optimal policy would not be able to solve the task. Second, this approximation implies that neither agent or the adversary can react to the action of the other and must choose simultaneously. This assumption is common in prior works \cite{bansal2017hamilton}. The order of the minimization and maximization is interchangeable. For both cases the optimal action as well as optimal model disturbance are identical and require the same approximation during the derivation.  
\hspace{\fill}\mbox{\qed}

\medskip
\subsection{Observation Disturbance Proof}
\begin{theorem*}
For the adversarial observation disturbance (\Eqref{eq:dyn_adv_obs}) with bounded signal energy (\Eqref{eq:signal_energy}), smooth drift and control matrix (i.e., $\va, \mB \in C^{1}$) and $\mB(\vx + \vxi_{o}) \approx \mB(\vx)$, the optimal continuous time policy $\pi^{k}$ and observation disturbance~$\vxi^{k}_{o}$ is described by
\begin{gather*}
\pi^{k}(\vx) = \nabla \tilde{g} \left(\mB(\vx)^{T} \nabla_x V^{k} \right) \hspace{30pt}
\vxi^{k}_{o} = - \alpha \frac{\vz_{o}}{\| \vz_{o} \|_2} \\
\text{with} \hspace{5pt} \vz_{o} = \left( \frac{\partial \va(\vx; \: \theta)}{\partial \vx} + \frac{\partial \mB(\vx; \: \theta)}{\partial \vx} \vu^{*} \right)^T V^{*}_{x}.
\end{gather*}
\end{theorem*}

\begin{table*}[t]
\tiny
\centering
\renewcommand{\arraystretch}{1.1}
\caption{\footnotesize
Average rewards on the simulated and physical systems. The ranking describes the decrease in reward compared to the best result averaged on all systems. 
The initial state distribution during training is noted by~$\mu$. The dynamics are either deterministic model $\theta \sim \delta(\theta)$ or sampled using uniform domain randomization $\theta \sim \mathcal{U}(\theta)$.
During evaluation the roll outs start with the pendulum pointing downwards. \vspace{-10pt}
}
\setlength{\tabcolsep}{6.2pt}
\begin{tabular*}{\textwidth}{l c c c c c c c c | c c  c c | c}
\toprule
 & & & \multicolumn{2}{c}{\textbf{Simulated Pendulum}}  & \multicolumn{2}{c}{\textbf{Simulated Cartpole}} & \multicolumn{2}{c|}{\textbf{Simulated Furuta Pendulum}}   & \multicolumn{2}{c}{\textbf{Physical Cartpole}} & \multicolumn{2}{c|}{\textbf{Physical Furuta Pendulum}} & \textbf{Average} \\
& & & 
Success & Reward & Success & Reward & Success & Reward  & Success & Reward & Success & Reward  & \textbf{Ranking}
\\  
 \multicolumn{1}{c}{Algorithm} & $\mu$ & $\theta$ &   [$\%$] & [$\mu \pm 2 \sigma$] & [$\%$] & [$\mu \pm 2 \sigma$] & [$\%$] & [$\mu \pm 2 \sigma$] & [$\%$] & [$\mu \pm 2 \sigma$] & [$\%$] & [$\mu \pm 2 \sigma$] & [$\%$] \\
 \cmidrule(lr){1-3} \cmidrule(lr){4-5} \cmidrule(lr){6-7} \cmidrule(lr){8-9} \cmidrule(lr){10-11} \cmidrule(lr){12-13} \cmidrule(lr){14-14} 
DP rFVI (ours) & $-$ & $\delta(\theta)$ 
& $100.0$ & $-032.7 \pm 000.3$ 
& $100.0$ & $-027.1 \pm 004.8$ 
& $100.0$ & $-041.3 \pm 010.8$ 
& $100.0$ &\textcolor{black}{$\mathbf{-074.1 \pm 040.3}$}
& $100.0$ & $-278.0 \pm 034.3$ 
& \textcolor{black}{$-062.7$}
\\
DP cFVI & $-$ & $\delta(\theta)$ 
& $100.0$ &\textcolor{black}{$\mathbf{-030.5 \pm 000.8}$}
& $100.0$ &\textcolor{black}{$\mathbf{-024.2 \pm 002.1}$}
& $100.0$ &\textcolor{black}{$\mathbf{-027.7 \pm 001.6}$}
& $73.3$ & $-143.7 \pm 210.4$ 
& $100.0$ &\textcolor{black}{$\mathbf{-082.1 \pm 007.6}$}
& \textcolor{black}{$-019.2$}
\\
RTDP cFVI (ours) & $\mathcal{U}$ & $\delta(\theta)$ 
& $100.0$ &\textcolor{black}{$\mathbf{-031.1 \pm 001.4}$}
& $100.0$ &\textcolor{black}{$\mathbf{-024.9 \pm 001.6}$}
& $100.0$ & $-040.1 \pm 002.7$ 
& $100.0$ &\textcolor{black}{$-101.1 \pm 029.0$}
& $00.0$ & $-1009.9 \pm 004.5$ 
& \textcolor{black}{$-247.7$}
\\
\cmidrule(lr){1-3} \cmidrule(lr){4-5} \cmidrule(lr){6-7} \cmidrule(lr){8-9} \cmidrule(lr){10-11} \cmidrule(lr){12-13} \cmidrule(lr){14-14}
SAC & $\mathcal{N}$ & $\mathcal{U}(\theta)$ 
& $100.0$ &\textcolor{black}{$\mathbf{-031.1 \pm 000.1}$}
& $100.0$ & $-026.9 \pm 003.2$ 
& $100.0$ & $-029.3 \pm 001.5$ 
& $00.0$ & $-518.6 \pm 028.1$ 
& $86.7$ & $-330.7 \pm 799.0$ 
& \textcolor{black}{$-185.8$}
\\
SAC \& UDR & $\mathcal{N}$ & $\delta(\theta))$ 
& $100.0$ & $-032.9 \pm 000.6$ 
& $100.0$ & $-029.7 \pm 004.6$ 
& $100.0$ & $-032.0 \pm 001.1$ 
& $100.0$ & $-394.8 \pm 382.8$ 
& $100.0$ & $-181.4 \pm 157.9$ 
& \textcolor{black}{$-120.8$}
\\
SAC & $\mathcal{U}$ & $\mathcal{U}(\theta)$ 
& $100.0$ &\textcolor{black}{$\mathbf{-030.6 \pm 001.4}$}
& $100.0$ &\textcolor{black}{$\mathbf{-024.2 \pm 001.4}$}
& $100.0$ &\textcolor{black}{$\mathbf{-028.1 \pm 002.0}$}
& $53.3$ & $-144.5 \pm 204.0$ 
& $100.0$ & $-350.8 \pm 433.3$ 
& \textcolor{black}{$-086.5$}
\\
SAC \& UDR & $\mathcal{U}$ & $\mathcal{U}(\theta)$ 
& $100.0$ & $-031.4 \pm 002.5$ 
& $100.0$ &\textcolor{black}{$\mathbf{-024.2 \pm 001.3}$}
& $100.0$ &\textcolor{black}{$\mathbf{-028.1 \pm 001.3}$}
& $40.0$ & $-296.4 \pm 418.9$ 
& $100.0$ & $-092.3 \pm 064.1$ 
& \textcolor{black}{$-063.8$}
\\
\cmidrule(lr){1-3} \cmidrule(lr){4-5} \cmidrule(lr){6-7} \cmidrule(lr){8-9} \cmidrule(lr){10-11} \cmidrule(lr){12-13} \cmidrule(lr){14-14}
DDPG & $\mathcal{N}$ & $\mathcal{U}(\theta)$ 
& $100.0$ &\textcolor{black}{$\mathbf{-031.1 \pm 000.4}$}
& $98.0$ & $-050.4 \pm 285.6$ 
& $100.0$ & $-030.5 \pm 003.5$ 
& $06.7$ & $-536.7 \pm 262.7$ 
& $46.7$ & $-614.1 \pm 597.8$ 
& \textcolor{black}{$-281.4$}
\\
DDPG \& UDR & $\mathcal{N}$ & $\delta(\theta))$ 
& $100.0$ & $-032.5 \pm 000.5$ 
& $100.0$ & $-027.4 \pm 002.3$ 
& $100.0$ & $-034.6 \pm 009.8$ 
& $00.0$ & $-517.9 \pm 117.6$ 
& $86.7$ & $-192.7 \pm 404.8$ 
& \textcolor{black}{$-156.6$}
\\
DDPG & $\mathcal{U}$ & $\mathcal{U}(\theta)$ 
& $100.0$ & $-031.5 \pm 000.7$ 
& $100.0$ & $-028.2 \pm 005.5$ 
& $100.0$ & $-030.0 \pm 001.7$ 
& $06.7$ & $-459.4 \pm 248.3$ 
& $100.0$ & $-146.6 \pm 218.3$ 
& \textcolor{black}{$-126.0$}
\\
DDPG \& UDR & $\mathcal{U}$ & $\mathcal{U}(\theta)$ 
& $100.0$ & $-032.5 \pm 003.6$ 
& $100.0$ & $-027.2 \pm 001.0$ 
& $100.0$ & $-032.1 \pm 001.5$ 
& $00.0$ & $-318.1 \pm 063.4$ 
& $100.0$ & $-156.7 \pm 246.4$ 
& \textcolor{black}{$-091.7$}
\\
\cmidrule(lr){1-3} \cmidrule(lr){4-5} \cmidrule(lr){6-7} \cmidrule(lr){8-9} \cmidrule(lr){10-11} \cmidrule(lr){12-13} \cmidrule(lr){14-14}
PPO & $\mathcal{N}$ & $\mathcal{U}(\theta)$ 
& $100.0$ & $-032.0 \pm 000.2$ 
& $100.0$ & $-031.5 \pm 007.2$ 
& $100.0$ & $-081.1 \pm 018.3$ 
& $00.0$ & $-287.9 \pm 068.8$ 
& $33.3$ & $-718.7 \pm 456.1$ 
& \textcolor{black}{$-261.7$}
\\
PPO \& UDR & $\mathcal{N}$ & $\delta(\theta))$ 
& $100.0$ & $-032.3 \pm 000.6$ 
& $100.0$ & $-084.0 \pm 007.8$ 
& $100.0$ & $-040.9 \pm 004.6$ 
& $00.0$ & $-435.4 \pm 111.9$ 
& $46.7$ & $-935.7 \pm 711.6$ 
& \textcolor{black}{$-370.0$}
\\
PPO & $\mathcal{U}$ & $\mathcal{U}(\theta)$ 
& $100.0$ & $-033.4 \pm 004.7$ 
& $99.0$ & $-039.7 \pm 045.7$ 
& $100.0$ & $-038.2 \pm 013.1$ 
& $00.0$ & $-183.8 \pm 018.0$ 
& $60.0$ & $-755.3 \pm 811.0$ 
& \textcolor{black}{$-219.4$}
\\
PPO \& UDR & $\mathcal{U}$ & $\mathcal{U}(\theta)$ 
& $100.0$ & $-035.6 \pm 003.1$ 
& $100.0$ & $-044.8 \pm 021.4$ 
& $100.0$ & $-048.5 \pm 006.2$ 
& $40.0$ & $-143.8 \pm 016.1$ 
& $100.0$ & \textcolor{black}{$\mathbf{-080.6 \pm 010.8}$}
& \textcolor{black}{$-054.4$}
\\
\bottomrule
\end{tabular*}
\vspace{-2.5em}
\label{table:cFVI_results}
\end{table*}

\medskip
\noindent
\emph{Proof.} \Eqref{eq:val_adv_update} can be written with the explicit constraint instead of the infimum with the admissible set $\Omega_{A}$, i.e., 
\begin{gather*}
V_{\text{tar}} = \max_{\vu} \: \min_{\vxi_{o}} \: r(\vx, \vu) + \gamma V\big(f(\vx, \vu, \: \vxi_{o})\big) \\ 
\text{with} \hspace{5pt} \frac{1}{2}\left((\vxi_{o}^T \vxi_{o} -\alpha^{2}\right) \leq \mathbf{0}.
\end{gather*}
Substituting the Taylor expansion for $V(\vx_{t+1})$, the dynamics models and reward as well as abbreviating $\mB(\vx + \vxi_{o}; \theta)$ as $\mB_{\xi}$ yields
\begin{align*}
V_{\text{tar}} &= \max_{\vu} \: \min_{\vxi_{o}} \: r + \gamma V + \gamma \nabla_x V^{T} f_c \Delta t + \gamma \mathcal{O}(.) \Delta t \\
\frac{V_\text{tar} - \gamma V}{\Delta t} &= q_c + \max_{\vu} \min_{\vxi} \left[\gamma \nabla_x V^{T} \left(\va_{\xi} + \mB_{\xi} \vu \right) + \gamma \mathcal{O}(.) - g_c \right]
\end{align*}
In the continuous time limit the optimal action and disturbance is determined by 
\begin{align*}
\vu^{*}, \vxi^{*}_{o} = \max_{\vu} \min_{\vxi} \left[\nabla_x V^{T} \left(\va_{\xi} + \mB_{\xi} \vu \right) - g_c(\vu) \right].
\end{align*}
This nested max-min optimization can be solved by first solving the inner optimization w.r.t. to $\xi$ and substituting this solution into the outer maximization. The Lagrangian for the optimal model disturbance is described by
\begin{align*}
\vxi^{*}_{o} = \argmin_{\vxi_{o}} \: \nabla_x V^{T} \big(\va(\vxi_{o}) + \mB(\vxi_{o}) \vu \big) + \frac{\lambda}{2} \left(\vxi_{o}^{T} \vxi_{o} - \alpha^{2}\right)
\end{align*}
Using the KKT conditions this optimization can be solved. The stationarity condition yields
\begin{gather*}
\vz_{o} + \lambda \: \vxi_{o}  = 0  \hspace{15pt} \Rightarrow \hspace{15pt} \vxi^{*}_{o} = - \frac{1}{\lambda} \vz_{o} \\
\text{with} \hspace{5pt} \vz_{o} =  \left[ \frac{\partial \va(\vx; \: \theta)}{\partial \vx} + \frac{\partial \mB(\vx; \: \theta)}{\partial \vx} \vu^{*} \right]^{T} \nabla_x V.
\end{gather*}
The primal feasibility and the complementary slackness yield
\begin{align*}
\frac{1}{2} \left(\frac{1}{\lambda^{2}} \vz_{o}^{T}\vz_{o} - \alpha^{2} \right) \leq 0 \hspace{10pt} &\Rightarrow \hspace{10pt} 
\lambda  \geq \frac{1}{\alpha} \| \vz_{\theta} \|_{2}  \\
\frac{\lambda}{2}  \left(\frac{1}{\lambda^{2}} \vz_{o}^{T} \vz_{o}  - \alpha^{2} \right) = 0 \hspace{10pt} &\Rightarrow \hspace{10pt} \lambda_{0} = 0, \hspace{5pt} \lambda_1 = \frac{1}{\alpha} \| \vz_{o} \|_{2}.
\end{align*}
Therefore, the optimal observation disturbance is described by 
\begin{align*}
\vxi^{*}_{o}(\vu) = -\alpha \frac{\vz_{o}}{\| \vz_{o} \|_2}.
\end{align*}
Then the optimal action can be computed by 
\begin{align*}
\vu^{*} = \argmax_{\vu} \nabla_x V^{T} \left[\va\big(\vxi^{*}_{o}(\vu)\big) + \mB\left(\vxi^{*}_{o}(\vu)\right) \vu \right] - g_c(\vu).
\end{align*}
Due to the envelope theorem, the extrema is described by
\begin{align*}
\mB(\vx; \theta + \vxi^{*}_{o}(\vu))^{T} \nabla_x V - g_{c}(\vu) = 0.
\end{align*}
This expression cannot be solved without approximation as $\mB$ does not necessarily be invertible w.r.t. $\vx$. Approximating $\mB(\vx + \vxi^{*}_{o}(\vu); \: \theta) \approx \mB(\vx; \theta)$, lets one solve for $\vu$. In this case the optimal action $\vu^{*}$ is described by  
$\vu^{*} = \nabla \tilde{g}(\mB(\vx; \theta)^{T} \nabla_x V)$. This approximation is feasible for two reasons. First of all, if the adversary can significantly alter the dynamics in each step, the system would not be controllable and the optimal policy would not be able to solve the task. Second, this approximation implies that neither agent or the adversary can react to the action of the other and must choose simultaneously. This assumption is common in prior works \cite{bansal2017hamilton}. The order of the minimization and maximization is interchangeable. For both cases the optimal action as well as optimal model disturbance are identical and require the same approximation during the derivation. \hspace{\fill}\mbox{\qed}

\section*{Value Function Representation}
\noindent
For the value function we are using a locally quadratic deep network as described in \cite{lutter2021cfvi}. This architecture assumes that the state cost is a negative distance measure between $\vx_t$ and the desired state $\vx_{\text{des}}$. Hence, $q_c$ is negative definite, i.e., $q(\vx) < 0 \:\: \forall \:\: \vx \neq \vx_{\text{des}}$ and $q(\vx_{\text{des}}) = 0$. 
These properties imply that $V^{*}$ is a negative Lyapunov function, as $V^{*}$ is negative definite, $V^{*}(\vx_{\text{des}}) = 0$ and $\nabla_{x}V^{*}(\vx_{\text{des}}) = \mathbf{0}$ \cite{khalil2002nonlinear}. With a deep network a similar representation can be achieved by
\begin{align*}
    V(\vx; \: \psi) &= -\left(\vx -  \vx_{\text{des}}\right)^T \mL(\vx;\:\psi) \mL(\vx;\:\psi)^T \left(\vx -  \vx_{\text{des}}\right) 
\end{align*}
with $\mL$ being a lower triangular matrix with positive diagonal. This positive diagonal ensures that $\mL \mL^T$ is positive definite. Simply applying a ReLu activation to the last layer of a deep network is not sufficient as this would also zero the actions for the positive values and $\nabla_{x}V^{*}(\vx_{\text{des}}) = \mathbf{0}$ cannot be guaranteed. The local quadratic representation guarantees that the gradient and hence, the action, is zero at the desired state. However, this representation can also not guarantee that the value function has only a single extrema at~$\vx_{\text{des}}$ as required by the Lyapunov theory. In practice, the local regularization of the quadratic structure to avoid high curvature approximations is sufficient as the global structure is defined by the value function target. $\mL$ is the mean of a deep network ensemble with $N$ independent parameters $\psi_i$. The ensemble mean smoothes the initial value function and is differentiable. Similar representations have been used by prior works in the safe reinforcement learning community \cite{berkenkamp2017safe, richards2018lyapunov, kolter2019learning, chang2019neural}. 

\begin{figure*}[t]
    \centering
    \includegraphics[width=\textwidth]{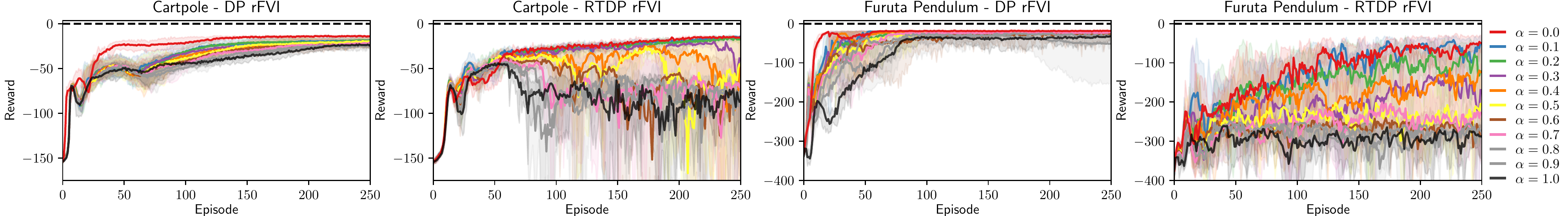}
    \vspace{-2.em}
    \caption{The learning curves for DP rFVI and RTDP rFVI with different adversary amplitudes averaged over $5$ seeds. The shaded area displays the \emph{min/max} range between seeds. The $\alpha$ corresponds of the percentage of the admissible set for all adversaries, i.e., with increasing $\alpha$ the adversary becomes more powerful. For DP rFVI the stronger adversaries do affect the final performance only marginally. For RTDP rFVI the adversaries become too powerful for small $\alpha$ and prevent learning of the optimal policy. This effect is especially distinct for the Furuta pendulum as this system is very sensible due to the low masses. Therefore, DP rFVI can learn a good optimal policy despite very strong adversaries.  
    } 
    \label{fig:ablation_learning_curves}\vspace{-0.8em}
\end{figure*} 

\section{Detailed Experimental Setup}
\medskip\noindent
\textbf{Systems} The performance of the algorithms is evaluated using the \emph{swing-up} the torque-limited pendulum, cartpole and Furuta pendulum. The physical cartpole (Figure \ref{fig:img_cartpole}) and Furuta pendulum (Figure \ref{fig:img_furuta}) are manufactured by Quanser~\cite{quanser}. For simulation, we use the equations of motion and physical parameters of the supplier. Both systems have very different characteristics. The Furuta pendulum consists of a small and light pendulum ($24$g, $12.9$cm) with a strong direct-drive motor. Even minor differences in the action cause large changes in acceleration due to the large amplification of the mass-matrix inverse. Therefore, the main source of uncertainty for this system is the uncertainty of the model parameters. The cartpole has a longer and heavier pendulum ($127$g, $33.6$cm). The cart is actuated by a geared cogwheel drive. Due to the larger masses the cartpole is not so sensitive to the model parameters. The main source of uncertainty for this system is the friction and the backlash of the linear actuator. The systems are simulated and observed with $500$Hz. The control frequency varies between algorithm and is treated as hyperparameter.

\medskip\noindent
\textbf{Reward Function} The desired state for all tasks is the upward pointing pendulum at $\vx_{\text{des}} = \mathbf{0}$. The state reward is described by $q_c(\vx) = -(\vz - \vz_{\text{des}})^T \mQ (\vz - \vz_{\text{des}})$ with the positive definite matrix~$\mQ$ and the transformed state $\vz$. For continuous joints the joint state is transformed to $z_i = \pi^{2} \sin(x_i)$. The action cost is described by $g_{c}(\vu) = - 2 \: \bm{\beta} \: \vu_{\text{max}} / \pi \log \cos(\pi \: \vu / (2 \: \vu_{\text{max}}))$ with the actuation limit~$\vu_{\text{max}}$ and the positive constant $\bm{\beta}$. This barrier shaped cost bounds the optimal actions. The corresponding policy is shaped by~$\nabla\tilde{g}(\vw) = 2\: \vu_{\text{max}} / \pi \: \tan^{-1}(\vw / \bm{\beta})$. For the experiments, the reward parameters are
\begin{align*}
    \text{Pendulum:}& &\mQ_{\text{diag}} &= \left[\textcolor{white}{0}1.0, \: 0.1 \right], & \beta &= 0.5\\
    \text{Cartpole:}& &\mQ_{\text{diag}} &= \left[25.0, \:1.0,\: 0.5, \: 0.1 \right], & \beta &= 0.1\\
    \text{Furuta Pendulum:}& & \mQ_{\text{diag}} &= \left[\textcolor{white}{0}1.0, \: 5.0, \: 0.1, \: 0.1 \right], & \beta &= 0.1
\end{align*}

\medskip
\section*{Additional Experimental Results}
\noindent
This section summarizes the additional experiments omitted in the main paper. In the following we perform an ablation study varying the admissible set and report the performance of additional baselines on the nominal physical system. 

\subsection*{Ablation Study - Admissible Set} \noindent
The ablation study highlighting the differences in learning curves for different admissible sets is shown in Figure \ref{fig:ablation_learning_curves}. In this plot we vary the admissible set of each adversary from $0$ to the $1$, where $\alpha = 1$ corresponds to the admissible set used for the robustness experiments. With increasing admissible set the final performance of the adversary decreases marginally for DP rFVI. For RTDP rFVI the optimal policy does not learn the task for stronger adversaries. RTDP rFVI starts to fail for $\alpha > 0.3$ on the cartpole. On the Furuta pendulum, RTDP rFVI starts to fail at $\alpha > 0.1$. The Furuta pendulum starts to fail earlier as this system is much more sensitive and smaller actions cause large changes in dynamics compared to the cartpole. This ablation study shows that the dynamic programming variant can learn the optimal policy despite strong adversaries. In contrast the real-time dynamic programming variant fails to learn the optimal policy for comparable admissible set. This failure is caused by the missing positive feedback during exploration. The adversary prevents the policy from discovering the positive reward during exploration. Therefore, the policy converges to a too pessimistic policy. In contrast the dynamic programming variant does not rely on exploration and covers the compact state domain. Therefore, the optimal policy discovers the optimal policy despite the strong adversary.

\subsection{Physical Experiments}
\noindent The rewards of all baselines on the nominal physical system are reported in Table \ref{table:cFVI_results}. The robustness experiments were only performed for the best performing baselines. On the nominal system rFVI outperforms all baselines on the cartpole. The domain randomization baselines do not necessarily outperform the deterministic deep RL baselines as the main source of simulation gap is the backslash and stiction of the actuator which is not randomized. For the Furuta pendulum, DP rFVI has a lower reward compared to DP cFVI. However, this lower reward is only caused by the chattering during the balancing that causes very high actions costs. If one only considers the swing-up phase, DP rFVI outperforms both PPO-U UDR and DP cFVI. For the Furuta pendulum the deep RL \& UDR baselines outperform the deep RL baselines without UDR. This is expected as the main uncertainty for the Furuta pendulum is caused by the uncertainty of the system parameters. In general a trend is observable that the algorithms with larger state domain during training achieve the better sim2real transfer.

\begin{figure*}[t]
    \centering
    \includegraphics[width=0.8\textwidth]{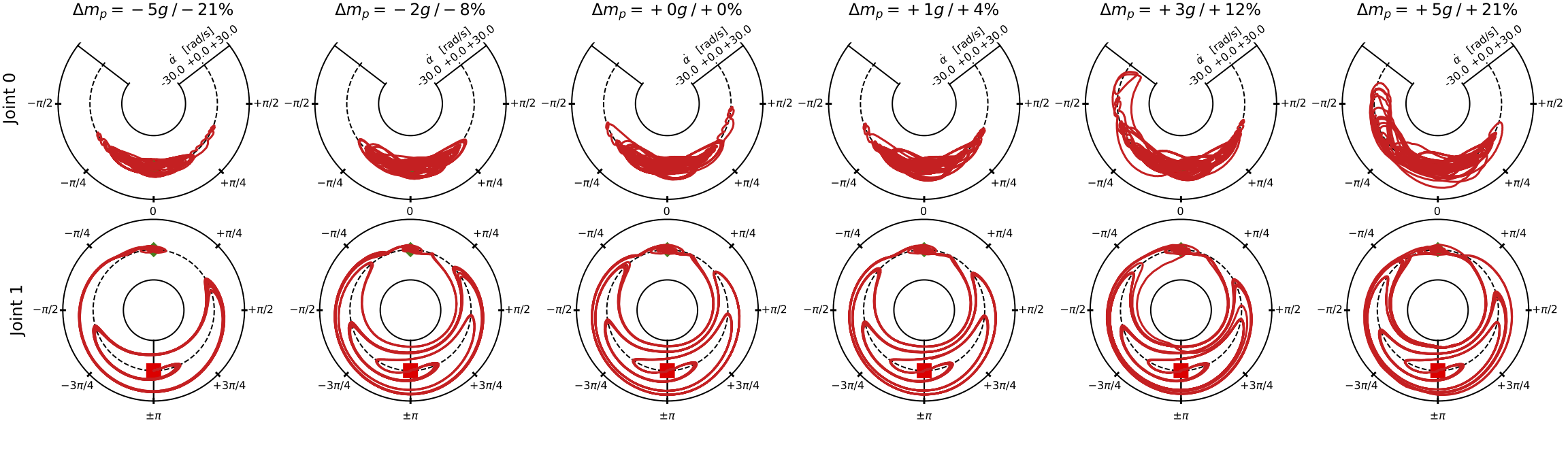}
    \includegraphics[width=0.8\textwidth]{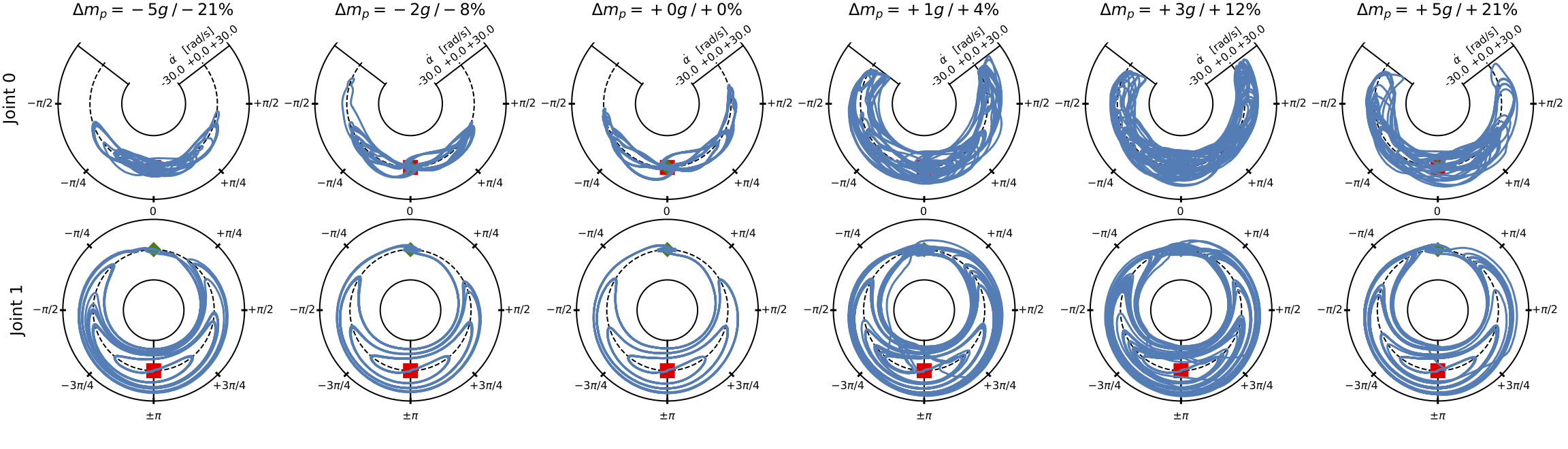}
    \includegraphics[width=0.8\textwidth]{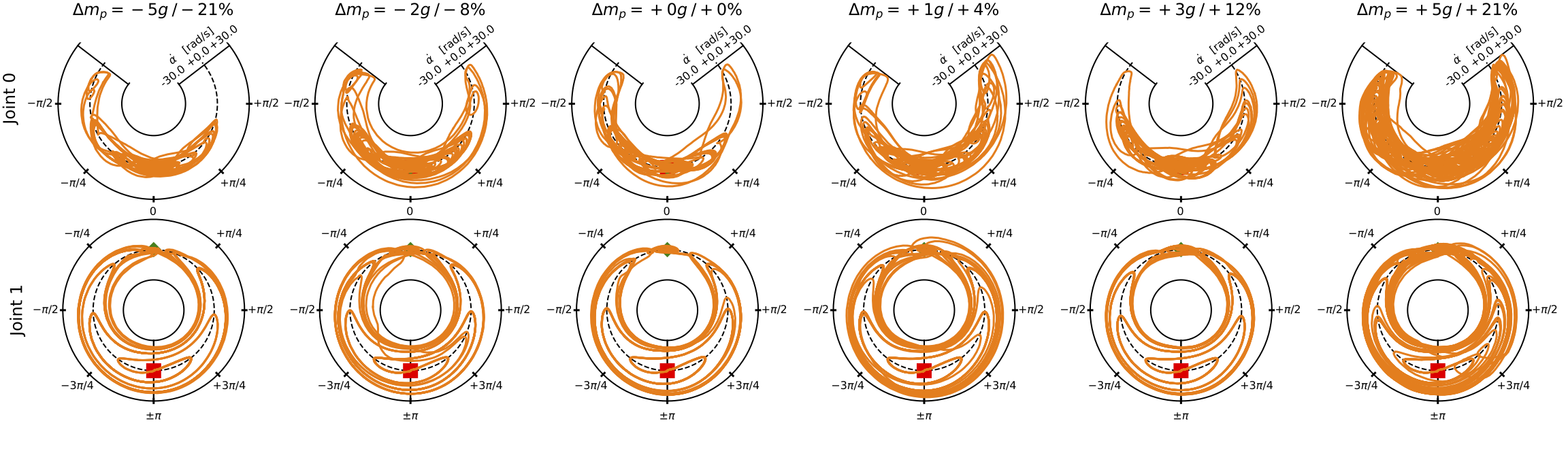}
    \includegraphics[width=0.8\textwidth]{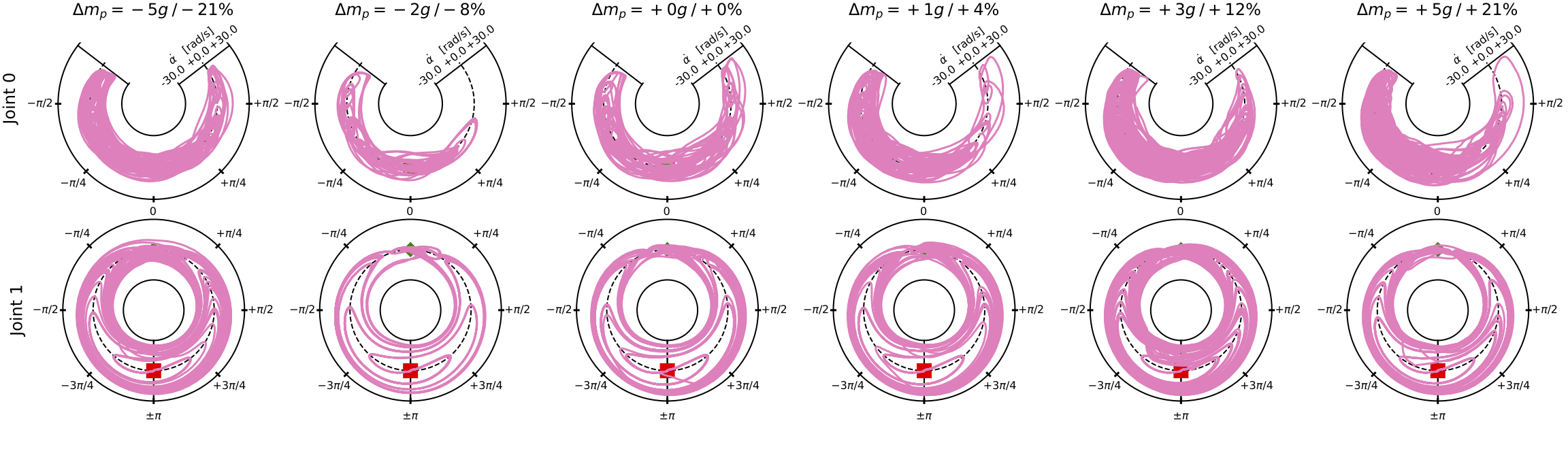}
    \includegraphics[width=0.8\textwidth]{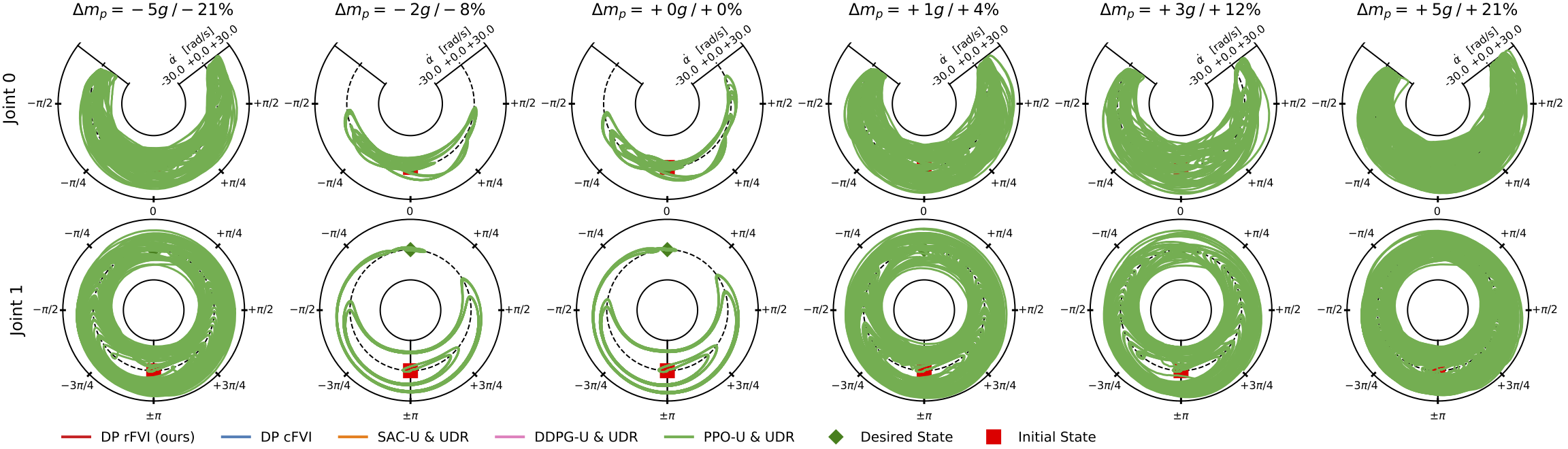}
    \caption{The roll-outs of DP rFVI, DP cFVI and the deep RL baselines with domain randomization the physical Furuta pendulum. The different columns correspond to different pendulum masses. The deviation from the dashed center line corresponds to the joint velocity. DP rFVI achieves a consistent swing-up for the different pendulum masses. In contrast to DP rFVI, the baselines start to deviate strongly from trajectories on the nominal system. When weights are added the baselines start to cover the complete state-space.}
    \label{fig:appendix_furuta_trajectories}
\end{figure*} 

\begin{figure*}[t]
    \centering
    \includegraphics[width=0.8\textwidth]{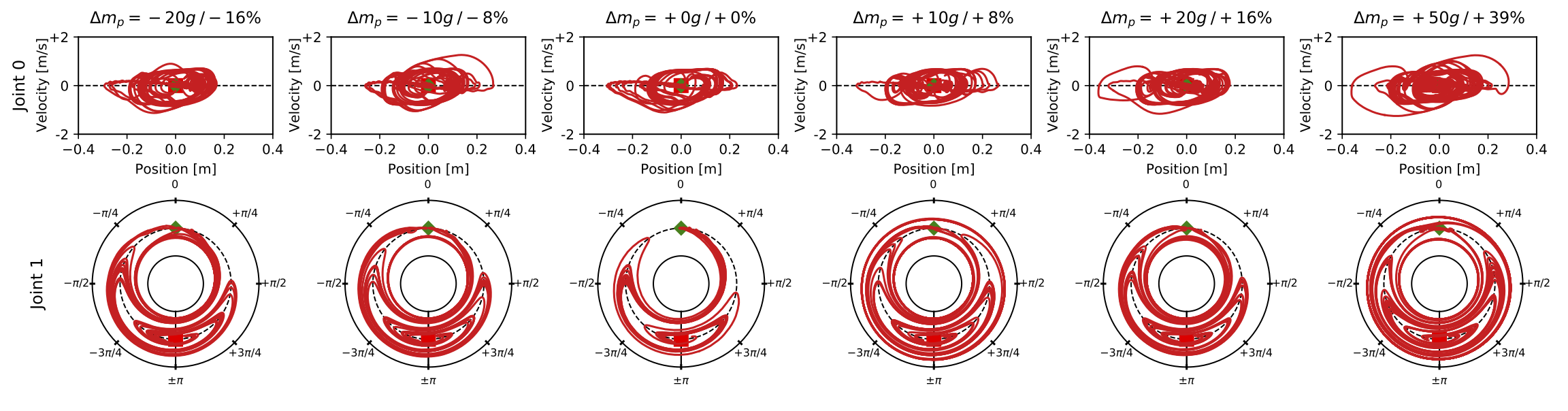}
    \includegraphics[width=0.8\textwidth]{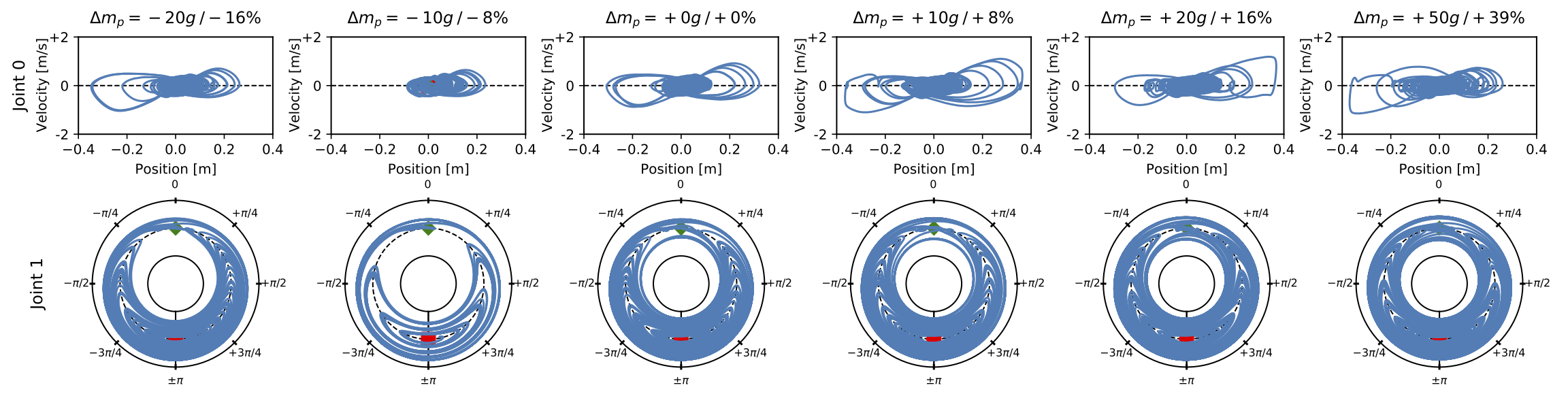}
    \includegraphics[width=0.8\textwidth]{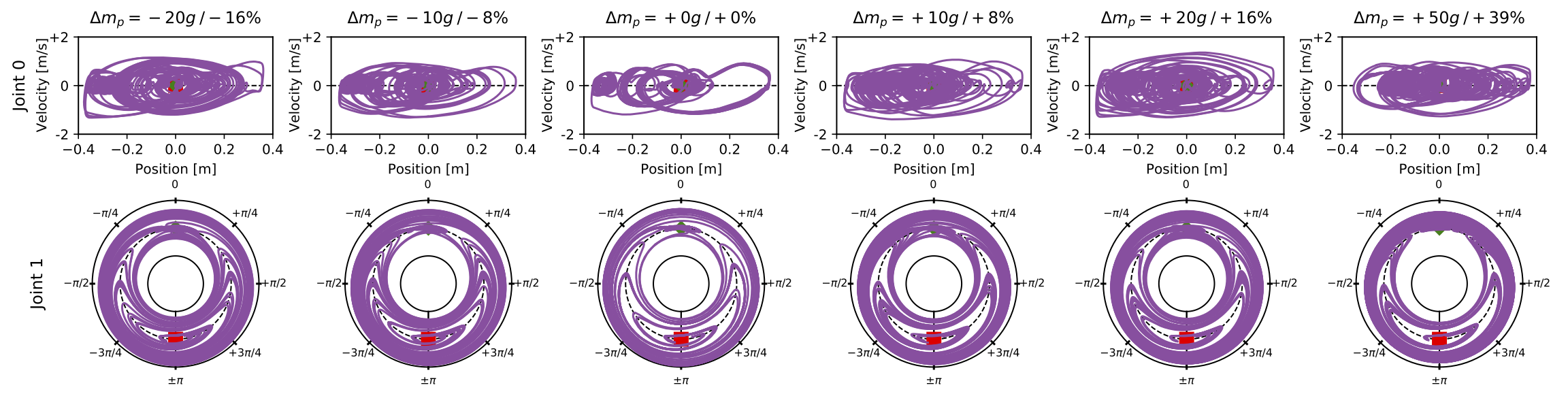}
    \includegraphics[width=0.8\textwidth]{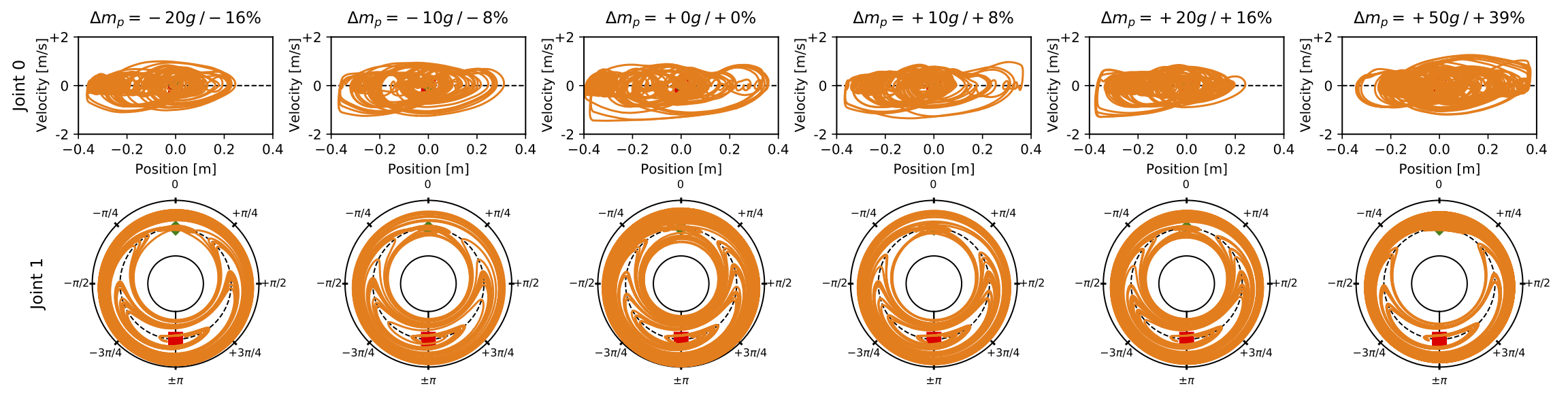}
    \includegraphics[width=0.8\textwidth]{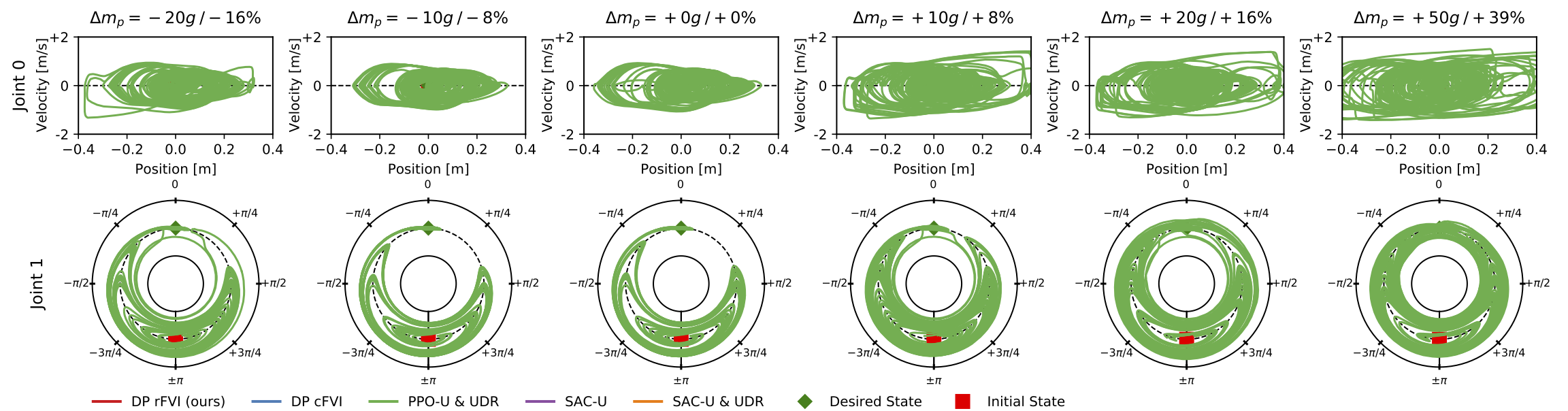}
    \caption{The roll-outs of DP rFVI, DP cFVI and the deep RL baselines with domain randomization the physical Cartpole. The different columns correspond to different pendulum masses. The deviation from the dashed center line corresponds to the joint velocity. DP rFVI achieves a consistent swing-up for the different pendulum masses but the failure rate slighlty increases when weights are added to the pendulum. In contrast to DP rFVI, the baselines start to deviate stronger from trajectories on the nominal system when the system dynamics are altered.}
    \label{fig:appendix_cartpole_trajectories}
\end{figure*} 
\end{document}